\documentclass[10pt,twocolumn,letterpaper]{article}
\usepackage[pagenumbers]{cvpr} 
\usepackage{graphicx}
\usepackage{amsmath}
\usepackage{amssymb}
\usepackage{booktabs}
\usepackage{multirow}
\usepackage{makecell}
\usepackage[table]{xcolor}
\usepackage[ruled,linesnumbered]{algorithm2e}
\usepackage{multirow}
\usepackage{diagbox}
\usepackage[pagebackref,breaklinks,colorlinks]{hyperref}

\usepackage[capitalize]{cleveref}
\crefname{section}{Sec.}{Secs.}
\Crefname{section}{Section}{Sections}
\Crefname{table}{Table}{Tables}
\crefname{table}{Tab.}{Tabs.}
\definecolor{mygray}{gray}{.88}

\usepackage{xspace}

\begin{document}

\title{Accelerating Dataset Distillation via Model Augmentation}

\author{
Lei Zhang\textsuperscript{1*} \hspace{2mm}
Jie Zhang\textsuperscript{1*} \hspace{2mm}
Bowen Lei\textsuperscript{2} \hspace{2mm} 
Subhabrata Mukherjee\textsuperscript{3} \hspace{2mm} \\
Xiang Pan\textsuperscript{4} \hspace{2mm}
Bo Zhao\textsuperscript{5} \hspace{2mm}
Caiwen Ding\textsuperscript{6} \hspace{2mm}
Yao Li\textsuperscript{7} \hspace{2mm}
Dongkuan Xu\textsuperscript{8$\dagger$} \\
\textsuperscript{1}Zhejiang University \hspace{3mm} \textsuperscript{2}Texas A\&M University \hspace{3mm} \textsuperscript{3}Microsoft Research \\ \textsuperscript{4}New York University \hspace{2mm} \textsuperscript{5}Beijing Academy of Artificial Intelligence \hspace{2mm} \textsuperscript{6}University of Connecticut \\ \textsuperscript{7}University of North Carolina, Chapel Hill \hspace{3mm} \textsuperscript{8}North Carolina State University  \\
{\tt\small \{zl\_leizhang, zj\_zhangjie\}@zju.edu.cn} \hspace{3mm}\tt\small dxu27@ncsu.edu
}

\newcommand{\customfootnotetext}[2]{{
\renewcommand{\thefootnote}{#1}
\footnotetext[0]{#2}}}

\maketitle

\begin{abstract}
Dataset Distillation (DD), a newly emerging field, aims at generating much smaller but efficient synthetic training datasets from large ones. 
Existing DD methods based on gradient matching achieve leading performance; however, they are extremely computationally intensive as they require continuously optimizing a dataset among thousands of randomly initialized models. In this paper, we assume that training the synthetic data with diverse models leads to better generalization performance. Thus we propose two \textbf{model augmentation} techniques, i.e. using \textbf{early-stage models} and \textbf{parameter perturbation} to learn an informative synthetic set with significantly reduced training cost. Extensive experiments demonstrate that our method achieves up to 20$\times$ speedup and comparable performance on par with state-of-the-art methods. 

\end{abstract}

\customfootnotetext{*}{Equal contribution.}
\customfootnotetext{${\dagger}$}{Corresponding author: Dongkuan Xu.}

\section{Introduction}
\label{sec:intro}

Dataset Distillation (DD)~\cite{DBLP:journals/corr/abs-1811-10959,DBLP:conf/cvpr/Cazenavette00EZ22b} or Dataset Condensation~\cite{DBLP:journals/corr/abs-2110-04181,DBLP:conf/icml/ZhaoB21}, aims to reduce the training cost by generating a small but informative synthetic set of training examples; such that the performance of a model trained on the small synthetic set is similar to that trained on the original, large-scale dataset.
Recently, DD has become an increasingly more popular research topic and has been explored in a variety of contexts, including federated learning~\cite{DBLP:journals/corr/abs-2204-01273,DBLP:journals/corr/abs-2208-11311}, continual learning~\cite{DBLP:conf/cvpr/MasarczykT20,DBLP:conf/ijcnn/SangermanoCCB22}, neural architecture search~\cite{DBLP:conf/icml/SuchRLSC20, DBLP:conf/iclr/ZhaoMB21}, medical computing~\cite{li2020soft, li2022compressed} and graph neural networks~\cite{DBLP:conf/iclr/JinZZLTS22,DBLP:journals/corr/abs-2206-13697}.

DD has been typically cast as a meta-learning problem~\cite{hospedales2021meta} involving bilevel optimization. For instance, Wang~\etal~\cite{DBLP:journals/corr/abs-1811-10959} formulate the network parameters as a function of the learnable synthetic set in the inner-loop optimization; then optimize the synthetic set by minimizing classification loss on the real data in the outer-loop. This recursive computation hinders its application to real-world large-scale model training, which involves thousands to millions of gradient descent steps. Several methods have been proposed to improve the DD method by introducing ridge regression loss \cite{DBLP:journals/corr/abs-2006-08572, DBLP:conf/iclr/NguyenCL21}, trajectory matching loss \cite{DBLP:conf/cvpr/Cazenavette00EZ22b}, \etc.
To avoid unrolling the recursive computation graph, Zhao~\etal~\cite{DBLP:conf/iclr/ZhaoMB21} propose to learn synthetic set by matching gradients generated by real and synthetic data when training deep networks. Based on this surrogate goal, several methods have been proposed to improve the informativeness or compatibility of synthetic datasets from other perspectives, ranging from data augmentation~\cite{DBLP:conf/icml/ZhaoB21}, contrastive signaling~\cite{DBLP:conf/icml/LeeCJYY22}, resolution reduction \cite{DBLP:conf/icml/KimKOYSJ0S22}, and bit encoding \cite{schirrmeister2022less}. 

\begin{figure}[t]
  \centering
   \includegraphics[width=1\linewidth]{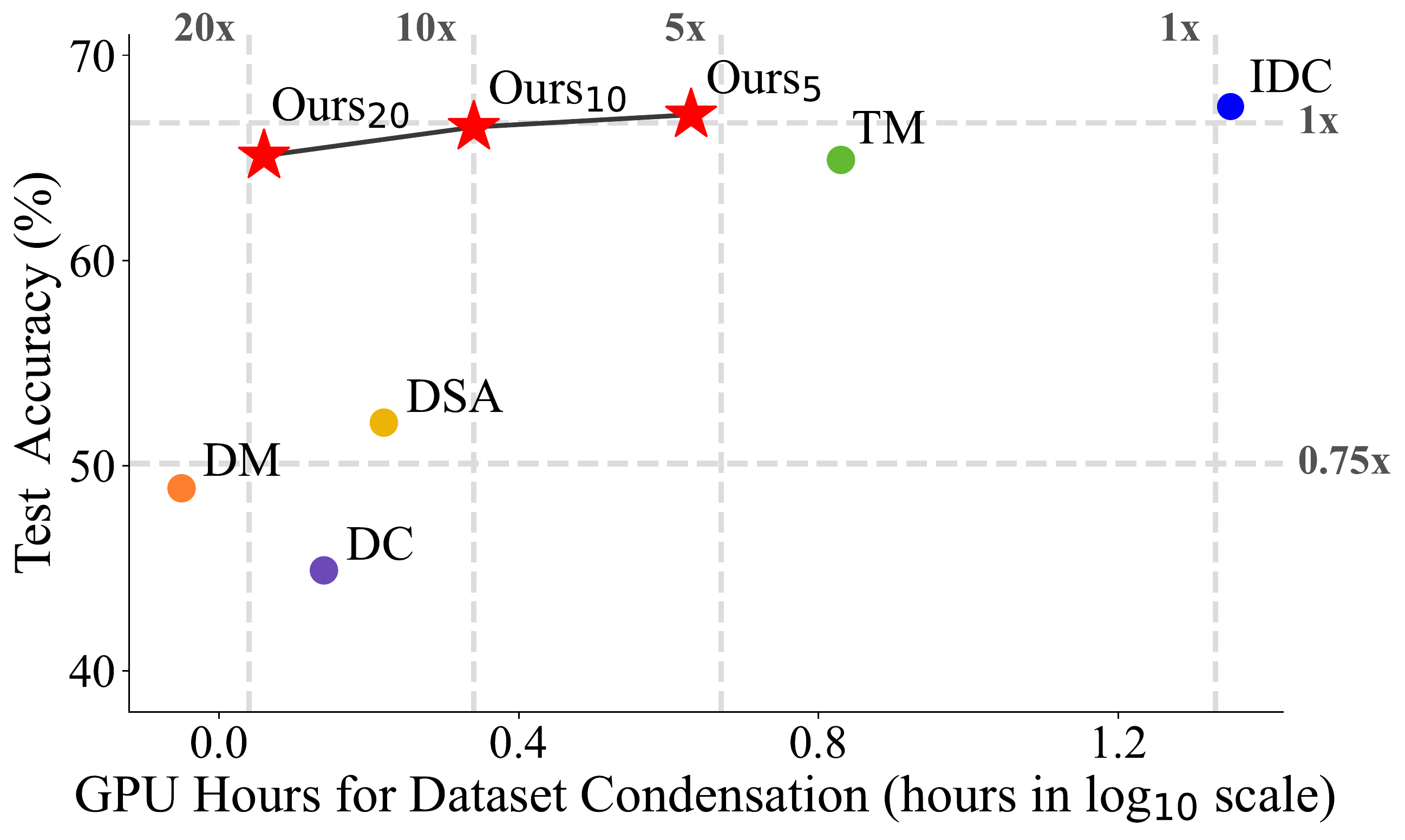}

   \caption{Performances of condensed datasets for training ConvNet-3 v.s. GPU hours to learn the  10 images per class condensed CIFAR-10 datasets with a single RTX-2080 GPU. $\text{Ours}_{5}$, $\text{Ours}_{10}$, and $\text{Ours}_{20}$ accelerates the training speed of the state-of-the-art method IDC~\cite{DBLP:conf/icml/KimKOYSJ0S22} $5\times$, $10\times$, and $20
   \times$ faster.
   }
   \label{fig:intro}
\end{figure}

Although model training on a small synthetic set is fast, the dataset distillation process is typically expensive. For instance, the state-of-the-art method IDC~\cite{DBLP:conf/icml/KimKOYSJ0S22} takes approximately 30 hours to condense 50,000 CIFAR-10 images into 500 synthetic images with a single RTX-2080 GPU, which is equivalent to the time it takes to train 60 ConvNet-3 models  
on the original dataset. Furthermore, the distillation time cost will rapidly increase for large-scale datasets \eg ImageNet-1K, which prevents its application in computation-limited environments like end-user devices. Prior work \cite{DBLP:journals/corr/abs-2110-04181} on reducing the distillation cost results in significant regression from the state-of-the-art performance.
In this paper, we aim to speed up the dataset distillation process, while preserving even improving the testing performance over state-of-the-art methods.

Prior works are computationally expensive as they focus on generalization ability such that the learned synthetic set is useful to train many different networks as opposed to a targeted network. This requires optimizing the synthetic set over thousands of differently initialized networks.
For example, IDC~\cite{DBLP:conf/icml/KimKOYSJ0S22} learns the synthetic set over 2000 randomly initialized models, while the trajectory matching method (TM)~\cite{DBLP:conf/cvpr/Cazenavette00EZ22b} optimizes the synthetic set for 10000 distillation steps with 200 pre-trained expert models.
Dataset distillation, which learns the synthetic data that is generalizable to unseen models, can be considered as an orthogonal approach to model training which learns model parameters that are generalizable to unseen data. Similarly, training the synthetic data with diverse models leads to better generalization performance.  
This intuitive idea leads to the following research questions:

\textbf{\textit{Question 1.}} \textit{How to design the candidate pool of models to learn synthetic data, for instance, consisting of randomly initialized, early-stage or well-trained models? }

Prior works \cite{DBLP:journals/corr/abs-1811-10959, DBLP:conf/iclr/ZhaoMB21, DBLP:conf/icml/KimKOYSJ0S22, DBLP:conf/cvpr/Cazenavette00EZ22b} use models from all training stages. The underlying assumption is that models from all training stages have similar importance. Zhao~\etal~\cite{DBLP:journals/corr/abs-2110-04181} show that synthetic sets with similar generalization performance can be learned with different model parameter distributions, given an objective function in the form of feature distribution matching between real and synthetic data. In this paper, we take a closer look at this problem and show that learning synthetic data on \textbf{early-stage} models is more efficient for gradient/parameter matching based dataset distillation methods.

\textbf{\textit{Question 2.}} \textit{Can we learn a good synthetic set using only a few models?
}

Our goal is to learn a synthetic set with a small number of (pre-trained) models to minimize the computational cost. However, using fewer models leads to poor generalization ability of the synthetic set. Therefore, we propose to apply \textbf{parameter perturbation} on selected early-stage models to incorporate model diversity and improve the generalization ability of the learned synthetic set. 

In a nutshell, we propose two \textbf{model augmentation} techniques to accelerate the training speed of dataset distillation, namely using \textbf{early-stage models} and \textbf{parameter perturbation} to learn an informative synthetic set with significantly less training cost. 
As illustrated in Fig.~\ref{fig:intro}., our method achieves up to 20$\times$ speedup and comparable performance on par with state-of-the-art DD methods.

\section{Related Work}
\label{sec:related}

\subsection{Dataset Distillation}

Recent advances in deep learning~\cite{he2022masked,he2020momentum,zhang2023delving,zhang2022towards,What_Transferred_Dong_CVPR2020,dong2022federated} rely on massive amounts of training data that not only consume a lot of computational resources, but it is also time-consuming to train these models on large data.
Dataset Distillation (DD) is introduced by Wang \etal~\cite{DBLP:journals/corr/abs-1811-10959}, in which network parameters are modeled as functions of synthetic data, and learned by gradient-based hyperparameter optimization~\cite{DBLP:conf/icml/MaclaurinDA15}. Subsequently, various works significantly improve the performance by learning on soft labels~\cite{DBLP:journals/corr/abs-2006-08572,DBLP:conf/ijcnn/SucholutskyS21}, optimizing via infinite-width kernel limit~\cite{DBLP:conf/iclr/NguyenCL21,DBLP:conf/nips/NguyenNXL21}, matching on gradient-space~\cite{DBLP:conf/iclr/ZhaoMB21,DBLP:journals/corr/abs-2208-00311}, model parameter-space~\cite{DBLP:conf/cvpr/Cazenavette00EZ22b}, and distribution space~\cite{DBLP:journals/corr/abs-2110-04181,DBLP:conf/cvpr/WangZPZYWHBWY22}, amplifying contrastive signals~\cite{DBLP:conf/icml/LeeCJYY22}, adopting data augmentations~\cite{DBLP:conf/icml/ZhaoB21}, and exploring regularity of dataset~\cite{DBLP:conf/icml/KimKOYSJ0S22}. DD has been 
applied to various scenarios including continual learning~\cite{DBLP:conf/cvpr/MasarczykT20,DBLP:conf/ijcnn/SangermanoCCB22,DBLP:journals/corr/abs-2103-15851}, privacy~\cite{DBLP:conf/icml/DongZL22}, federated learning~\cite{DBLP:journals/corr/abs-2008-04489,DBLP:journals/corr/abs-2204-01273,DBLP:journals/corr/abs-2207-09653}, graph neural network~\cite{DBLP:conf/kdd/JinTJLZTY22,DBLP:conf/iclr/JinZZLTS22}, neural architecture search~\cite{DBLP:conf/icml/SuchRLSC20} for images~\cite{DBLP:conf/cvpr/Cazenavette00EZ22}, text~\cite{DBLP:journals/corr/abs-2104-08448}, and medical imaging data~\cite{DBLP:journals/corr/abs-2209-14603}. In addition to the efforts made to improve performance and expand applications, few studies have focused on the efficiency of DD. This is a critical and practical problem closely related to the real-world application of DD.

\subsection{Efficient Dataset Distillation}

In this work, we focus on the efficiency of dataset distillation algorithm, which is under-explored in previous works. Zhao~\etal~\cite{DBLP:journals/corr/abs-2110-04181} make improvements in efficiency via distribution matching in random embedding spaces, which replaces expensive bi-level optimization in common methods~\cite{DBLP:conf/iclr/ZhaoMB21,DBLP:conf/icml/KimKOYSJ0S22}. However, the speed-up of DD in their work results in a significant drop in performance, which exhibits a large gap between their method and other SOTA DD methods~\cite{DBLP:conf/icml/KimKOYSJ0S22}. Cazenavette~\etal~\cite{DBLP:conf/cvpr/Cazenavette00EZ22} improve efficiency via parameter matching in pre-trained networks. However, they need to pre-train 100 networks from scratch on real data, which leads to massively increased computational resources. In this work, we seek to significantly reduce training time and lower computational resources, while maintaining comparable performance.

\section{Preliminary}
\label{sec:preliminary}


The goal of dataset distillation is to generate a synthetic dataset $\mathcal{S}$ from the original training dataset $\mathcal{T}$ such that an arbitrary model trained on $\mathcal{S}$ is similar to the one trained on $\mathcal{T}$. Among various dataset distillation approaches~\cite{DBLP:journals/corr/abs-2110-04181,DBLP:conf/cvpr/Cazenavette00EZ22b,DBLP:conf/icml/KimKOYSJ0S22, DBLP:conf/nips/NguyenNXL21}, gradient-matching methods have achieved state-of-the-art performance. However, they require a large amount of training time and expensive computational resources. In this paper, we propose to use gradient matching to reduce the computational requirement while maintaining similar performance.

\noindent \textbf{Gradient Matching.}
Gradient-matching dataset distillation approach~\cite{DBLP:conf/iclr/ZhaoMB21} matches the network gradients on synthetic dataset $\mathcal{S}$ to the gradients on real dataset $\mathcal{T}$. The overall training object can be formulated as:
\begin{equation}
\label{eq:gradient-matching}
\begin{aligned}
    &\underset{\mathcal{S}}{\operatorname{maximize}} \  \sum_{t=0}^{T} \operatorname{Cos}\left(\nabla_\theta \ell\left(\theta_t ; \mathcal{S}\right), \nabla_\theta \ell\left(\theta_t ; \mathcal{T}\right)\right) \\
    & \wrt \quad \theta_{t+1}=\theta_t-\eta \nabla_\theta \ell\left(\theta_t ; \mathcal{S}\right) \ \\
\end{aligned}
\end{equation}
where $\theta_{t}$ denotes the network weights at the $t^{\text{th}}$ training step from the randomly initialized weights $\theta_{0}$ given $\mathcal{S}$, $\ell(\theta, \mathcal{S})$ denotes the training loss for weight $\theta$ and the dataset $\mathcal{S}$, $\ell$ denotes loss function, and $\text{Cos}(\cdot, \cdot)$ denotes the channel-wise cosine similarity.

In addition, recent works have made various efforts to enhance the performance of gradient-matching from the perspective of data diversity. Zhao \etal~\cite{DBLP:conf/icml/ZhaoB21} utilize differentiable siamese augmentation to synthesize more informative images. Kim \etal~\cite{DBLP:conf/icml/KimKOYSJ0S22} explore the regularity of dataset to strengthen the representability of condensed datasets.

\noindent\textbf{Discussion on Efficiency.} Current works~\cite{DBLP:conf/icml/ZhaoB21, DBLP:conf/iclr/ZhaoMB21,DBLP:conf/icml/KimKOYSJ0S22} use a large number of randomly initialized networks (\eg, 2000) to improve the generalization performance of condensed dataset. The huge number of models makes the DD process time-consuming and computation-expensive. For instance, condensing 1 image per class in a synthetic dataset of CIFAR-10 by using state-of-the-art method IDC~\cite{DBLP:conf/icml/KimKOYSJ0S22} consumes 200k epochs of updating network, in addition to the 2,000k epochs of updating $\mathcal{S}$, which requires over 22.2 hours on a single RTX-2080 GPU. While Zhao~\etal~\cite{DBLP:journals/corr/abs-2110-04181} make efforts to solve computation the challenge by using distribution-matching instead of gradient-matching -- reducing number of updates from 200k to 20k and training time from 22.2 hours to 0.83 hours -- the accuracy of condensed data also degrades dramatically from $50.6\%$ to $26.0\%$. This potentially results from the redundant learning on randomly initialized networks.


\begin{figure}[t]
  \centering
   \includegraphics[width=1.\linewidth]{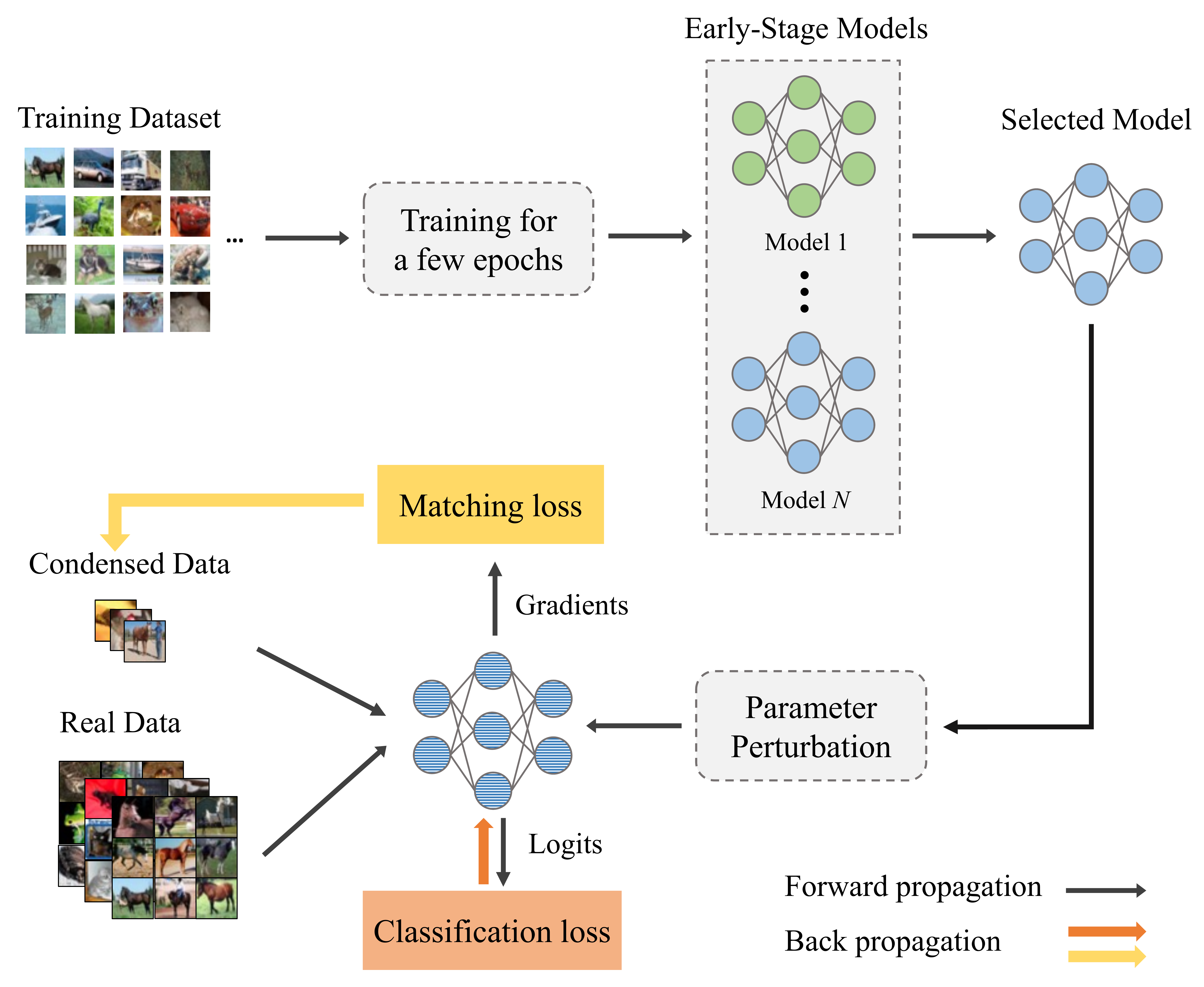}

   \caption{The illustration of our proposed fast dataset distillation method. We perform early-stage pretraining and parameter perturbation on models in dataset distillation.
   }
   \label{fig:scenario}
\end{figure}

\section{Method}
\label{sec:method}

\subsection{Overview}

We illustrate the framework of our proposed efficient dataset distillation method in \cref{fig:scenario}. Our method consists of three stages: 1) Early-stage Pre-training, 2) Parameter Perturbation, and 3) Distillation via gradient-matching. In stage 1, we utilize pre-trained networks at the early stage as an informative parameter space for dataset distillation. In stage 2, we conduct parameter perturbation on models selected from stage 1 to further augment the diversity of model parameter distribution. In stage 3, the synthetic dataset is optimized with gradient-matching strategy on {these augmented models from early stages}.

\subsection{Early-Stage Models:\ Initializing with Informative Parameter Space}
Existing gradient-matching methods~\cite{DBLP:conf/icml/ZhaoB21,DBLP:conf/iclr/ZhaoMB21,DBLP:conf/icml/KimKOYSJ0S22} train synthetic data on a large number of randomly initialized networks for learning to generalize to unseen initializations. Furthermore, the initialized networks will be updated for many SGD steps in the inner-loop for learning better synthetic data, which requires much computational resources.




Data augmentation is frequently used to prevent overfitting and improve generalization performance when optimizing deep networks~\cite{DBLP:conf/ijcai/Wen0YSGWX21, DBLP:conf/icml/WuZVR20}. Similarly, we propose to use model augmentation to improve the generalization performance when learning condensed datasets. 
Inspired by ModelSoups~\cite{DBLP:conf/iclr/LopesDC22,DBLP:conf/icml/WortsmanIGRLMNF22}, a practical method to improve performance of model ensembles, we pre-train a set of networks with different hyper-parameters, including learning rate, random seed, and data augmentation, so that we construct a parameter space with rich diversity.
Instead of leveraging randomly initialized networks in each outer loop in traditional methods, we sample those early-stage networks as the initialization, which are more informative for implementing gradient matching.

Comparing with well-trained networks, using early-stage networks have two benefits. First, early-stage networks require less training cost. Second, the early-stage networks have rich diversity~\cite{DBLP:journals/corr/abs-1812-04754,DBLP:conf/iclr/SagunEGDB18,DBLP:journals/corr/abs-1711-08856} and provide large gradients~\cite{DBLP:conf/iclr/FrankleSM20}, which leads to better gradient matching. More discussion can be found in the supplementary.

\subsection{Parameter Perturbation:\ Diversifying Parameter Space}



Motivated by the data perturbation which is widely used to diversify the training data for better knowledge distillation~\cite{DBLP:conf/nips/NamYLL21,DBLP:conf/icml/NamLH022}, we propose to conduct the model perturbation in dataset distillation for further diversifying the parameter space. We implement perturbation after sampling the network (parameters) from the early-stage parameter space in each outer loop.

We formulate our fast dataset distillation as the gradient-matching on parameter-perturbed early-stage models between real data and synthetic data:
\begin{equation}
\begin{aligned}
& \min_{\mathcal{S}} D\left(\nabla_\theta \ell\left(\hat{\theta} ; \mathcal{S}\right), \nabla_\theta \ell\left(\hat{\theta} ; \mathcal{T}\right)\right)\\
& \wrt \;\; \hat{\theta} \leftarrow \theta^{\mathcal{T}} + \alpha \cdot \mathbf{d}
\text {, }
\end{aligned}
\label{euqation:loss}
\end{equation}
where $\theta^{\mathcal{T}}$ represents network weights trained on real data $\mathcal{T}$, $D$ denotes a distance-based matching objective, and $\alpha$ is the magnitude of parameter perturbation. $\mathbf{d}$ is sampled from a Gaussian distribution $\mathcal{N}(0,\mathbf{I})$ with dimensions compatible with network parameter $\theta$ and filter normalized by 
\begin{equation}
\mathbf{d}_{l, j} \leftarrow \frac{\mathbf{d}_{l, j}}{\left\|\mathbf{d}_{l, j}\right\|_F + \epsilon}\left\|\theta_{l, j}\right\|_F
\end{equation}
to eliminate the scaling invariance of neural networks~\cite{DBLP:conf/nips/Li0TSG18}, where $\mathbf{d}_{l, j}$ is the $j$-th filter at the $l$-th layer of $\mathbf{d}$ and $\|\cdot\|_F$ denotes the Frobenius norm. $\epsilon$ is a small positive constant.

\begin{algorithm}[t]
\caption{Efficient Dataset Distillation}\label{alg:one}
\SetNoFillComment
\SetKwInOut{Output}{Notation}
\SetKwInput{KwResult}{Definition}
\SetKwData{Data}{Definition}
\KwIn{Training data $\mathcal{T}$, loss function $l$, number of\\ classes $C$, number of model $N$, magnitude $\alpha$, augmentation function $\mathcal{A}$, multi-information function $f$, deep neural network $\psi_{\theta}$ parameterized with $\theta$}
\KwOut{Condensed dataset $\mathcal{S}$}
\KwResult{$\left.D\left(B, B^{\prime} ; \theta\right)=\| \nabla_\theta \ell(\theta ; B)-\nabla_\theta \ell\left(\theta ; B^{\prime}\right) \|\right.$}
\tcc{Early-Stage Pre-train}
Randomly initialize $N$ networks $\{\tau_{1}...\tau_{N}\}$\;
\For {$n\gets1$ \KwTo $N$}{
Update network $\tau_{n}$ on real data $\mathcal{T}$:\\
\For {$p\gets1$ \KwTo $P$}{ 
    \quad ${\tau}_{n, p+1} \leftarrow {\tau}_{n,p}-\eta {\nabla}_{{\tau}_{n,p}} \ell\left({\tau}_{n,p} ; \mathcal{A}(\mathcal{T})\right)$ \\
}
}
Initialize condensed dataset $\mathcal{S}$\\
\For{$t\gets0$ \KwTo $T$}{
    Randomly load one checkpoint from $\left\{ \tau_{1}...\tau_{N} \right\}$ to initialize $\psi_{\theta}$ \;
    \tcc{Parameter Perturbation}
    Sample vector $\mathbf{d}$ from Gaussian distribution \\
    Parameter perturbation on $\psi_{\theta}$: $\theta \gets \theta + \alpha \cdot \mathbf{d}$ \\
    \For{$m\gets0$ \KwTo $M$} {
        \For{$c\gets0$ \KwTo $C$} {
            Sample an intra-class mini-batch $T_{c} \sim \mathcal{T}$, $S_{c} \sim \mathcal{S}$\\
            Update synthetic data $\mathcal{S}_{c}$:\\
            $S_c\leftarrow S_c-\lambda \nabla_{S_c} D\left(\mathcal{A}\left(f(S_c)\right), \mathcal{A}\left(T_c\right)\right)$ \\
        }
        Sample a mini-batch $T \sim \mathcal{T}$ \\
        Update network $\psi_{\theta}$ w.r.t classification loss: \\
        \quad $\theta_{m+1} \leftarrow \theta_m-\eta \nabla_\theta \ell\left(\theta_m ; \mathcal{A}(T)\right)$ \\
    }
}
\end{algorithm}

\subsection{Training Algorithm}

We depict our method in \cref{alg:one}. We build our training algorithm on the state-of-the-art method IDC~\cite{DBLP:conf/icml/KimKOYSJ0S22}. Before dataset distillation, we pre-trained $N$ models on real data for only a few epochs. This is significantly cheaper than existing methods that well-train many networks till convergence. We train the condensed dataset $\mathcal{S}$ for $T$ outer loops and $M$ inner loops. At each outer loop, we randomly select a model from $N$ early-stage models as initialization and employ parameter perturbation on it. At each inner loop, we optimize the synthetic samples $\mathcal{S}$ by minimizing the gradient matching loss with regard to the sampled real batch $\mathcal{T}_{c}$ and real synthetic batch $\mathcal{S}_{c}$ of the same class $c$, respectively. The network $\theta_{m}$ is then updated on real data. 
Please refer to \cite{DBLP:conf/icml/KimKOYSJ0S22} for more details.
The numbers of pre-train epochs $P$ and outer loop $K$ are relatively small. 
In experiments, we set $P=2$ compared with 300 for a well-trained network and $K=400$ compared with 2000 in SOTA DD method IDC~\cite{DBLP:conf/icml/KimKOYSJ0S22}. Note that our method can also be easily applied to other dataset distillation methods for reducing training time, and we explore it in \cref{sec:analysis}.

\section{Experiments}
\label{sec:experiment}

\begin{table*}[!ht]
    \centering
    \begin{tabular}{clrrrrr}
    \toprule
    \multirow{2}{*}{Dataset} & \multirow{2}{*}{Method} & \multicolumn{3}{c}{Img/Cls} & \multirow{2}{*}{\makecell{Speed Up}} & \multirow{2}{*}{\makecell{Acc. Gain}} \\
    \Xcline{3-5}{0.4pt}
    {} & {} & \multicolumn{1}{c}{1} & \multicolumn{1}{c}{10} & \multicolumn{1}{c}{50} & {} \\
    \midrule
    \multirow{8}{*}{CIFAR-10} & {Full Dataset} & {88.1} & {88.1} & {88.1} & {-} & {-}\\
    {} & {IDC~\cite{DBLP:conf/icml/KimKOYSJ0S22}} & {50.6 \ (21.7h)} & {67.5 \ (22.2h)} & {74.5 \ (29.4h)} & {$1.00\times$} & {$1.00\times$} \\
    {} & {CAFE~\cite{DBLP:conf/cvpr/WangZPZYWHBWY22}} & {30.3} & {46.3} & {55.5} & {-} &{$0.54\times$} \\
    {} & {DSA~\cite{DBLP:conf/icml/ZhaoB21}} & {28.2 \ (0.09h)} & {52.1 \ (1.94h)} & {60.6 \ (11.1h)} & {$85.0\times$} & {$0.71\times$} \\
    {} & {DM~\cite{DBLP:journals/corr/abs-2110-04181}} & {26.0 \ (0.25h)} & {48.9 \ (0.26h)} & {63.0 \ (0.31h)} & {$89.0\times$} & {$0.69\times$}\\
    {} & {TM~\cite{DBLP:conf/cvpr/Cazenavette00EZ22b}} & {46.3 \ (6.35h)} & {65.3 \ (6.69h)} & {71.6 \ (7.39h)} & {$3.57\times$} & {$0.94\times$} \\
    \rowcolor{mygray} {} & {$\text{Ours}_{\text{5}}$} & {49.2 \ (4.44h)} & {67.1 \ (4.45h)} & {73.8 \ (6.11h)} & {$\boldsymbol{4.90\times}$} & {$\boldsymbol{0.99\times}$} \\
    \rowcolor{mygray} {} & {$\text{Ours}_{\text{10}}$} & {48.5 \ (2.22h)} & {66.5 \ (2.23h)} & {73.1 \ (3.05h)} & {$\boldsymbol{9.77\times}$} & {$\boldsymbol{0.97\times}$} \\
    \midrule
    \multirow{9}{*}{CIFAR-100} & {Full Dataset} & {56.2} & {56.2} & {56.2} & {-} & {-}\\
    {} & {IDC~\cite{DBLP:conf/icml/KimKOYSJ0S22}} & {25.1 \ (125h)} & {45.1 \ (127h)} & {-} & {$1.00\times$} & {$1.00\times$} \\
    {} & {CAFE~\cite{DBLP:conf/cvpr/WangZPZYWHBWY22}} & {12.9} & {27.8} & {37.9} & {-} & {$0.56\times$} \\
    {} & {DSA~\cite{DBLP:conf/icml/ZhaoB21}} & {13.9 \ (0.83h)} & {32.3 \ (17.5h)} & {42.8 \ (221.1h)} & {$78.9\times$} & {$0.63\times$} \\
    {} & {DM~\cite{DBLP:journals/corr/abs-2110-04181}} & {11.4 \ (1.67h)} & {29.7 \ (2.64h)} & {43.6 \ (2.78h)} & {$61.4\times$} & {$0.55\times$} \\
    {} & {TM~\cite{DBLP:conf/cvpr/Cazenavette00EZ22b}} & {24.3 \ (7.74h)} & {40.1 \ (9.47h)} & {47.7 \ (-)} & {$14.7\times$} & {$0.92\times$} \\
    \rowcolor{mygray} {} & {$\text{Ours}_{\text{5}}$} & {29.8 \ (25.1h)} & {45.6 \ (25.6h)} & {52.6 \ (42.00h)} & {$\boldsymbol{4.97\times}$} & {$\boldsymbol{1.10\times}$} \\
    \rowcolor{mygray} {} & {$\text{Ours}_{\text{10}}$} & {29.4 \ (12.5h)} & {45.2 \ (12.8h)} & {52.2 \ (21.00h)} & {$\boldsymbol{9.96\times}$} & {$\boldsymbol{1.09\times}$} \\
    \rowcolor{mygray} {} & {$\text{Ours}_{\text{20}}$} & {29.1 \ (6.27h)} & {44.1 \ (6.40h)} & {52.1 \ (10.50h)} & {$\boldsymbol{19.9\times}$} & {$\boldsymbol{1.07\times}$} \\
    \bottomrule
    \end{tabular}
    \caption{Comparing efficiency and performance of dataset distillation methods on CIFAR-10 and CIFAR-100. Speed up represents the average acceleration amount of training time on a single RTX-2080 GPU with the same batch size 64. Acc. Gain represents the average improvement in test accuracy of network trained on the condensed dataset over IDC~\cite{DBLP:conf/icml/KimKOYSJ0S22}. Training time is not reported for CAFE~\cite{DBLP:conf/cvpr/WangZPZYWHBWY22} that does not provide official implementation and IDC~\cite{DBLP:conf/icml/KimKOYSJ0S22} that requires more than one GPU on CIFAR-100 for Img/Cls=50.}
    \label{tab:cifar}
    \vspace{2mm}
\end{table*}

In this section, we first evaluate our method on various datasets against state-of-the-art baselines. Next, we examine the proposed method in depth with ablation analysis.

\begin{table*}[!ht]
    \centering
    \begin{tabular}{clrrrr}
    \toprule
    \multirow{2}{*}{Dataset} & \multirow{2}{*}{Method} & \multicolumn{2}{c}{Img/Cls} & \multirow{2}{*}{\makecell{Speed Up}} & \multirow{2}{*}{\makecell{Acc. Gain}} \\
    \Xcline{3-4}{0.4pt}
    {} & {} &  \multicolumn{1}{c}{10} & \multicolumn{1}{c}{20} & {} \\
    \midrule
    \multirow{5}{*}{ImageNet-10} & {Full Dataset} & {90.8}  & {90.8} & {-} & {-} \\
    {} & {IDC~\cite{DBLP:conf/icml/KimKOYSJ0S22}} & {72.8 \ (70.14h)}  & {76.6 \ (92.78h)} & {$1.00\times$} & {$1.00\times$} \\
    {} & {DSA~\cite{DBLP:conf/icml/ZhaoB21}} & {52.7 \ (26.95h)} & {57.4 \ (51.39h)}  & {$2.20\times$} & {$0.73\times$} \\
    {} & {DM~\cite{DBLP:journals/corr/abs-2110-04181}} & {52.3 \ (1.39h)}  & {59.3 \ (3.61h)} & {$38.1\times$} & {$0.74\times$} \\
    \rowcolor{mygray} {} & {$\text{Ours}_{\text{5}}$} & {74.6 \ (15.52h)} & {76.3 \ (20.05h)}  & {$\boldsymbol{4.57\times}$} & {$\boldsymbol{1.01\times}$} \\
    \midrule
    \multirow{5}{*}{ImageNet-100} & {Full Dataset} & {82.0} & {82.0}  & {-} & {-} \\
    {} & {IDC~\cite{DBLP:conf/icml/KimKOYSJ0S22}} & {46.7 \ (141h)}  & {53.7 \ (185h)} & {$1.00\times$} & {$1.00\times$} \\
    {} & {DSA~\cite{DBLP:conf/icml/ZhaoB21}} & {21.8 (9.72h)} & {30.7 (23.9h)} & {$14.1\times$} & {$0.51\times$} \\
    {} & {DM~\cite{DBLP:journals/corr/abs-2110-04181}} & {22.3 (2.78h)} & {30.4 (2.81h)} & {$58.2\times$} & {$0.52\times$} \\
    \rowcolor{mygray} {} & {$\text{Ours}_{\text{5}}$} & { 48.4\ (29.8h)}  & { 56.0 \ (38.6h)} & {$\boldsymbol{4.76\times}$} & {$\boldsymbol{1.04\times}$} \\
    \bottomrule
    \end{tabular}
    \caption{Comparing efficiency and performance of dataset distillation methods on ImageNet-10 and ImageNet-100. We measure the training time on a single RTX-A6000 GPU with the same training hyperparameters. For ImageNet-100, we follow IDC~\cite{DBLP:conf/icml/KimKOYSJ0S22} to split the whole dataset into five tasks with 20 classes each for faster optimization. The training time reported in ImageNet-100 is for one task.}
    \label{tab:imagenet}
\end{table*}

\begin{figure*}
  \centering
  \begin{subfigure}{0.24\linewidth}
    \includegraphics[width=1.\linewidth]{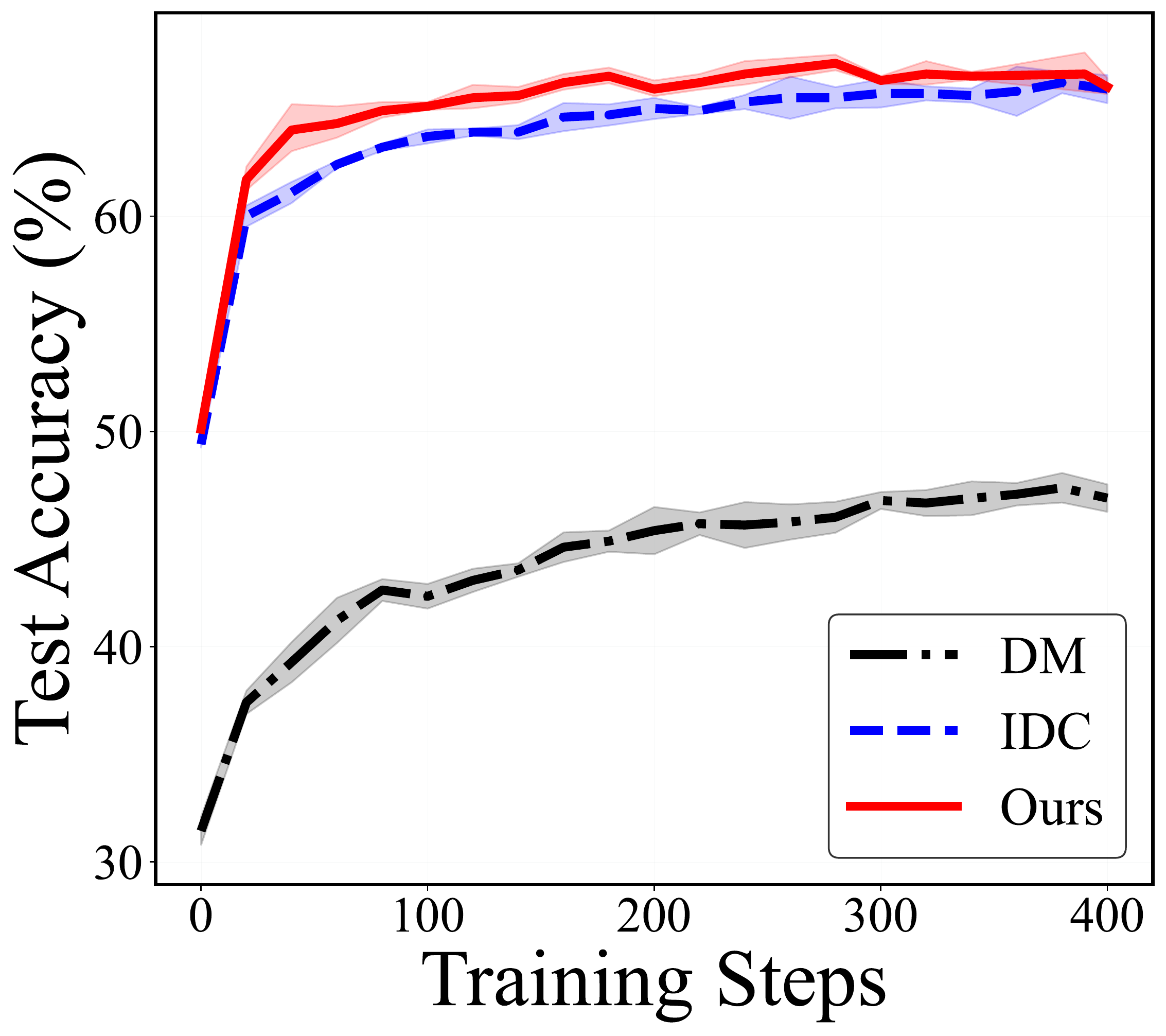}
    \caption{CIFAR10 (Img/Cls=10)}
    \label{fig:cifar10-10-epoch}
  \end{subfigure}
  \hfill
  \begin{subfigure}{0.24\linewidth}
    \includegraphics[width=1\linewidth]{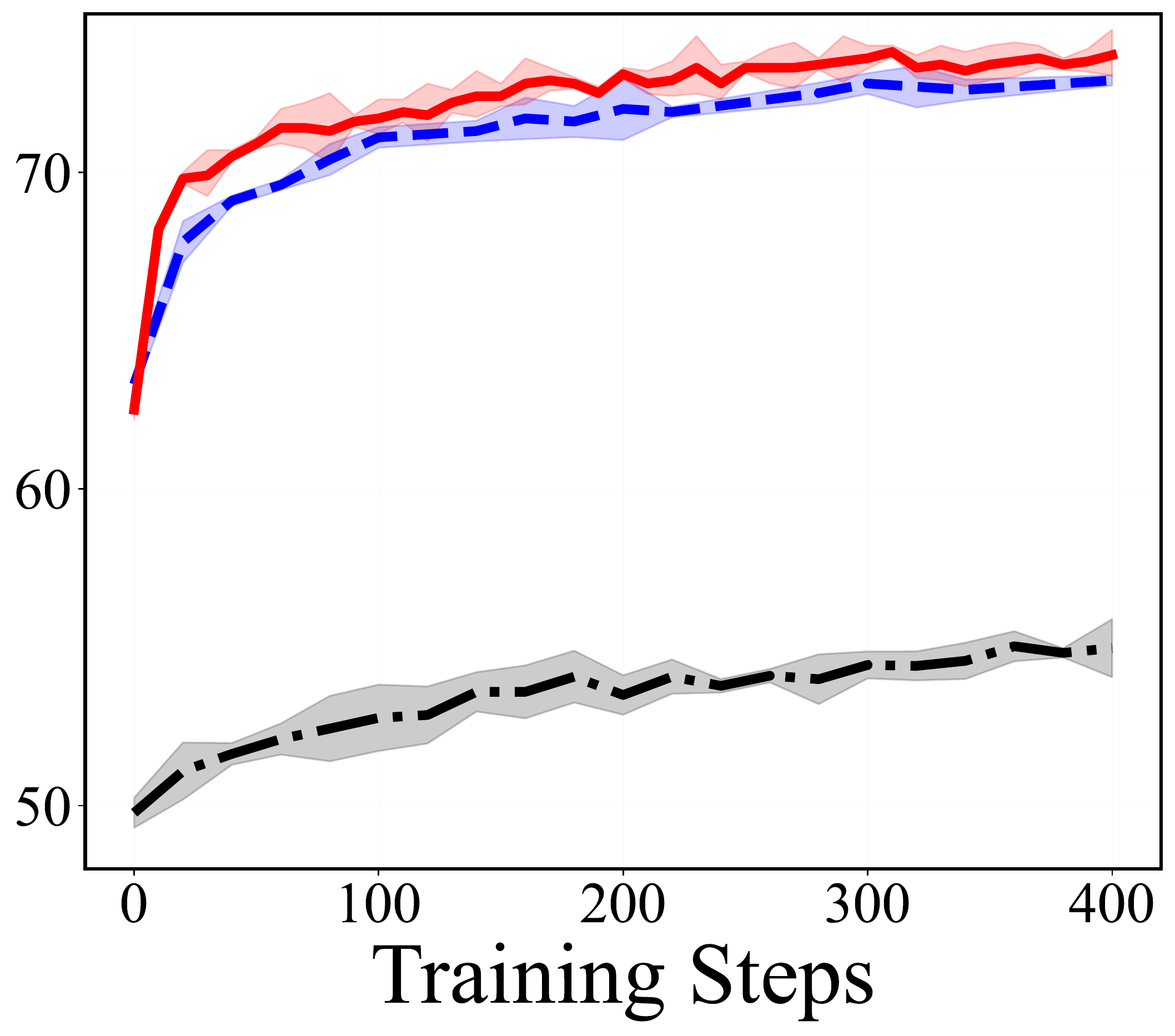}
    \caption{CIFAR10 (Img/Cls=50)}
    \label{fig:cifar10-50-epoch}
  \end{subfigure}
  \hfill
  \begin{subfigure}{0.24\linewidth}
    \includegraphics[width=1\linewidth]{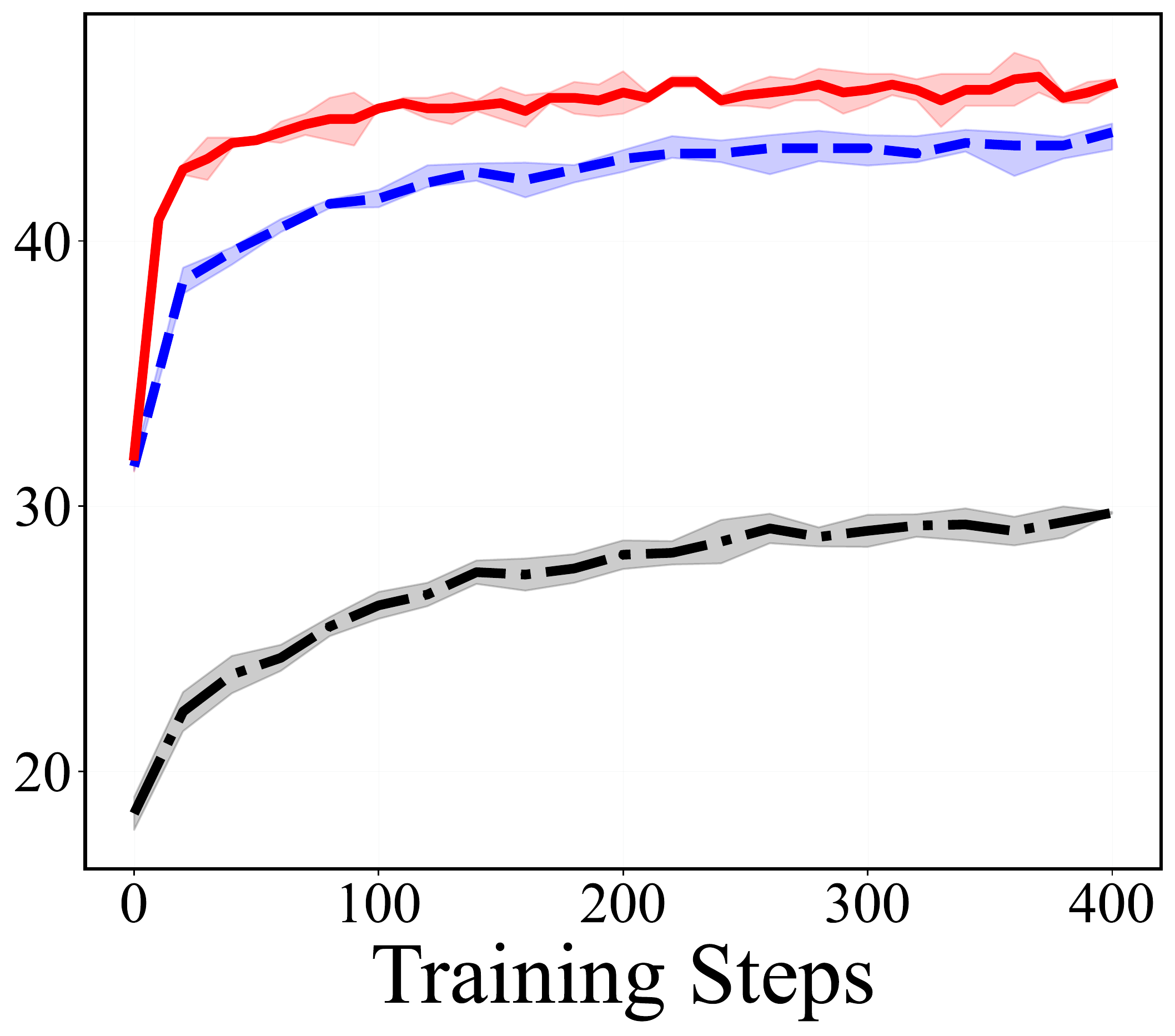}
    \caption{CIFAR100 (Img/Cls=10)}
    \label{fig:cifar100-10-epoch}
  \end{subfigure}
  \hfill
  \begin{subfigure}{0.24\linewidth}
    \includegraphics[width=1\linewidth]{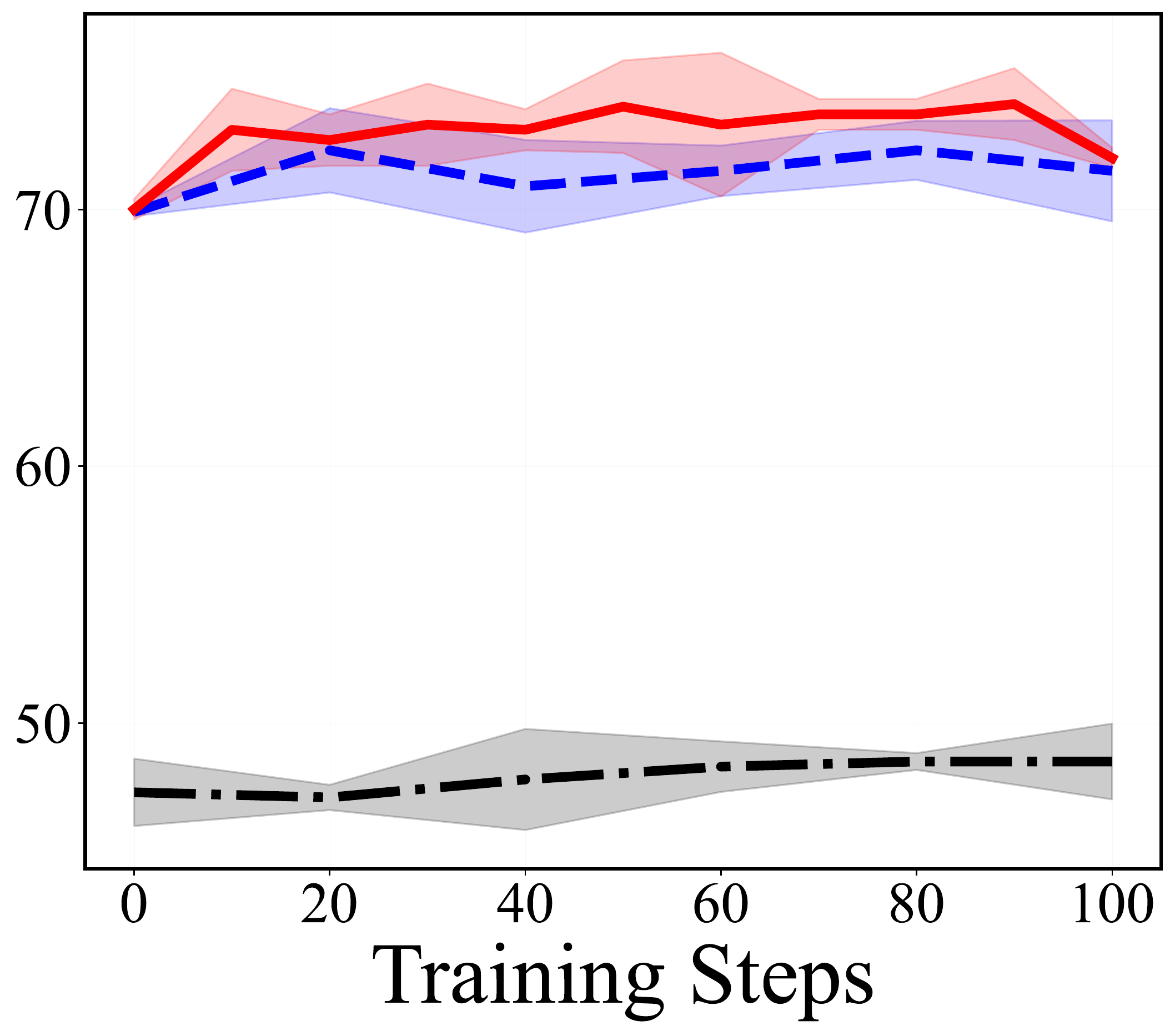}
    \caption{ImageNet-10 (Img/Cls=10)}
    \label{fig:imagenet10-10-epoch}
  \end{subfigure}
  \caption{Performance comparison across a varying number of training steps.}
  \label{fig:epoch-accuracy}
  \vspace{3.5mm}
\end{figure*}

\begin{figure*}
  \centering
  \begin{subfigure}{0.235\linewidth}
    \includegraphics[width=1\linewidth]{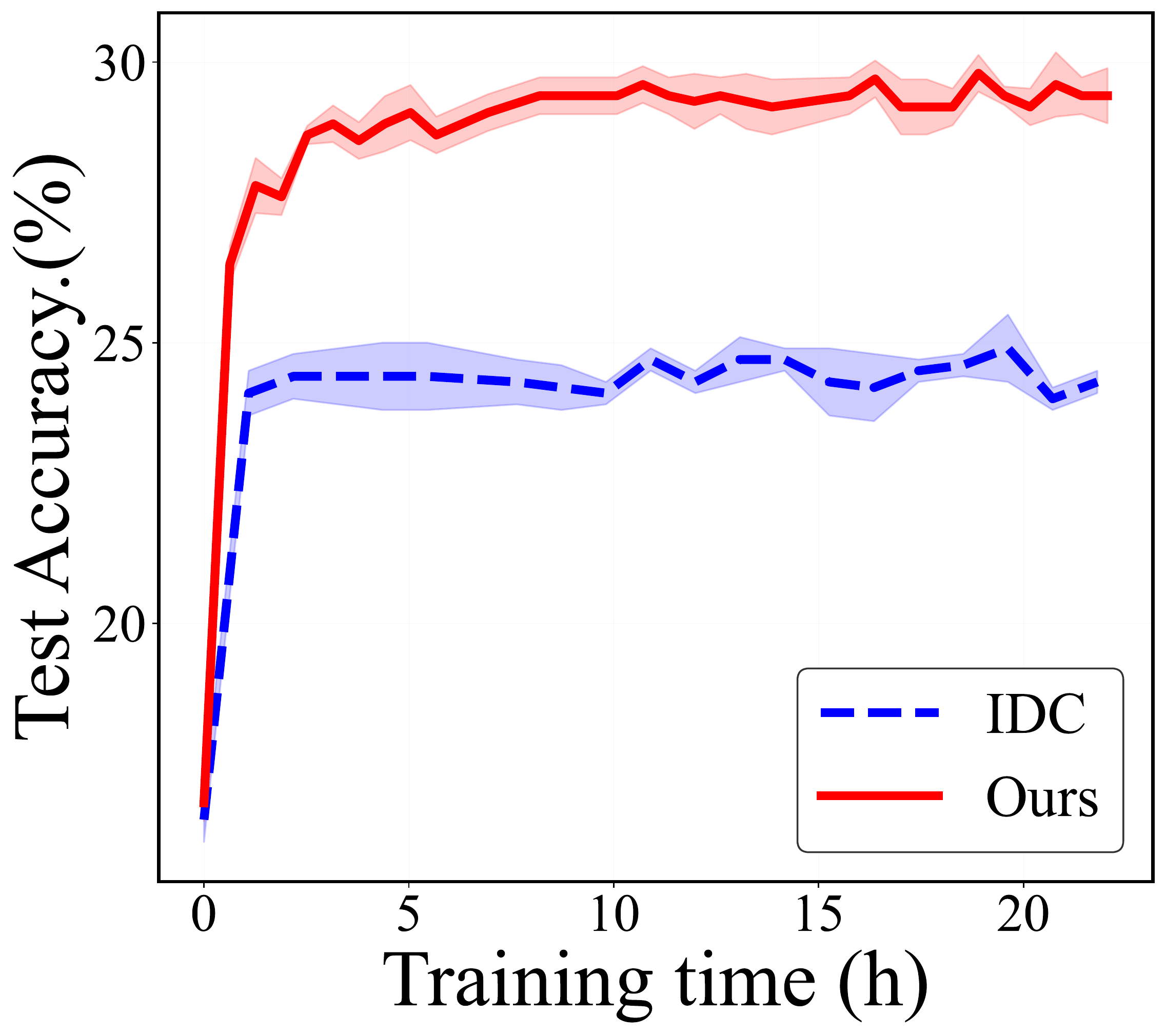}
    \caption{CIFAR-100 (Img/Cls=1).}
    \label{fig:cifar100-1-time}
  \end{subfigure}
  \hfill
  \begin{subfigure}{0.235\linewidth}
    \includegraphics[width=1.\linewidth]{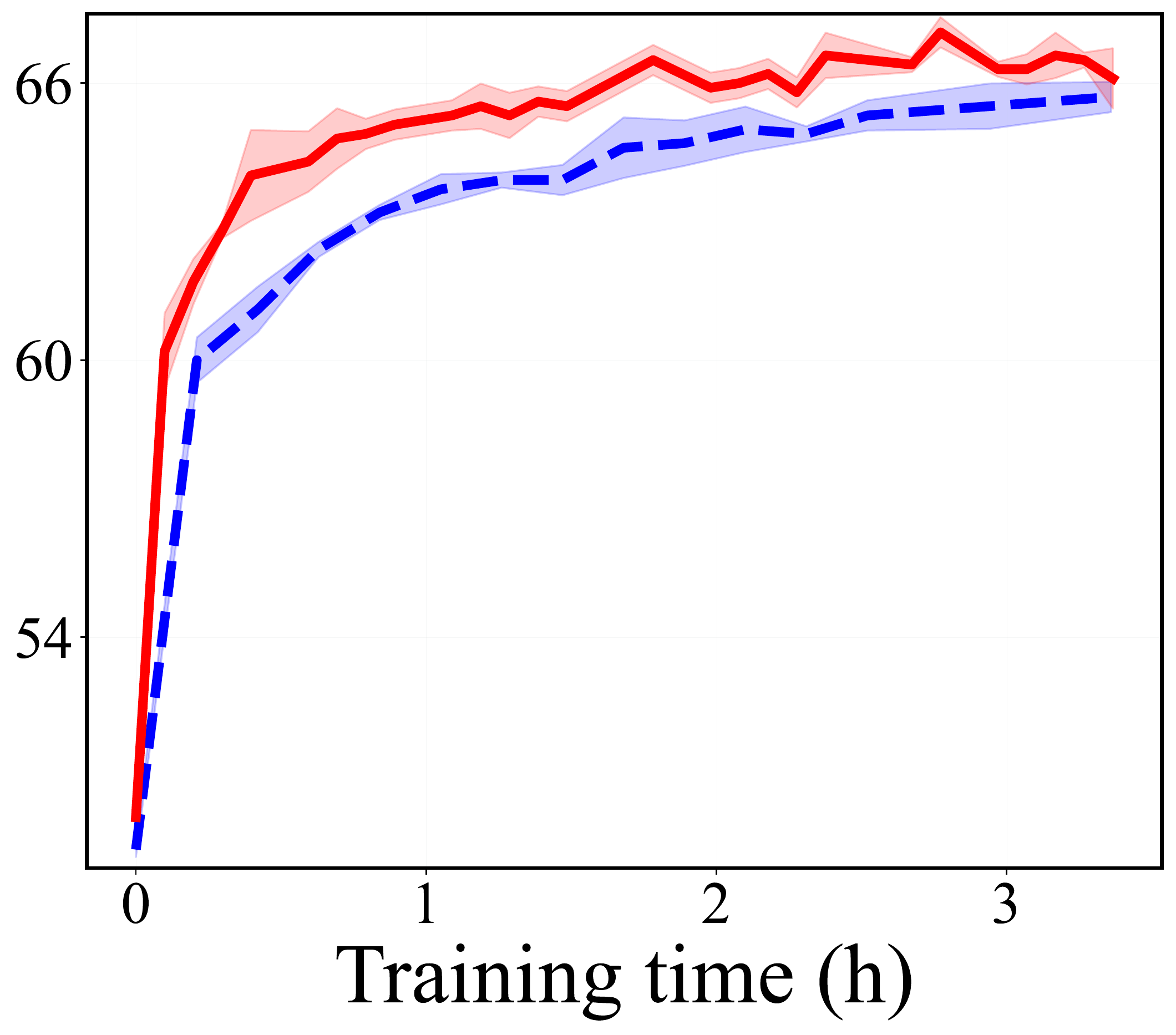}
    \caption{CIFAR-10 (Img/Cls=10)}
    \label{fig:imagenet10-20-time}
  \end{subfigure}
  \hfill
  \begin{subfigure}{0.235\linewidth}
    \includegraphics[width=1.\linewidth]{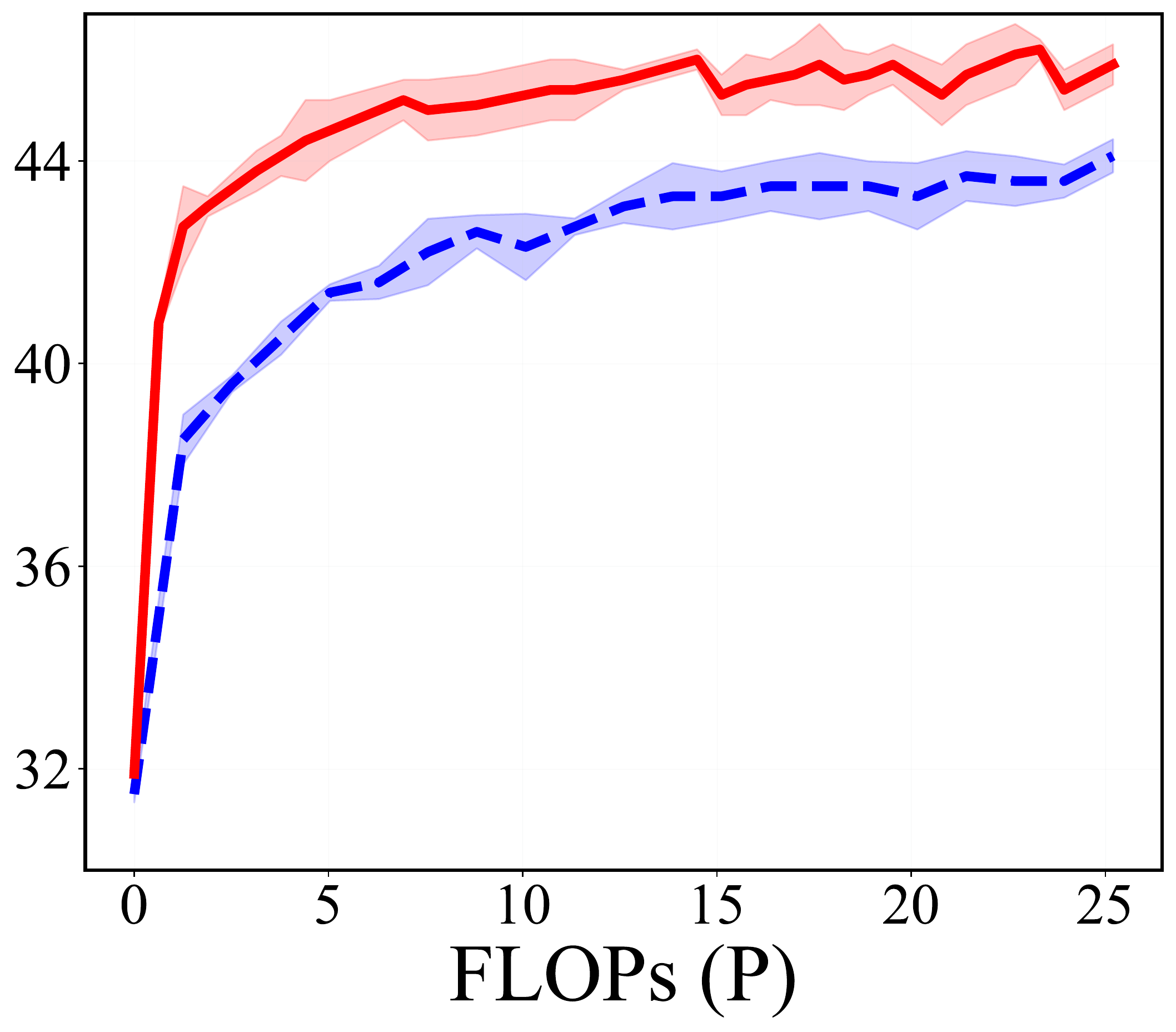}
    \caption{CIFAR100 (Img/Cls=10)}
    \label{fig:cifar100-10-flops}
  \end{subfigure}
  \hfill
  \begin{subfigure}{0.235\linewidth}
    \includegraphics[width=1\linewidth]{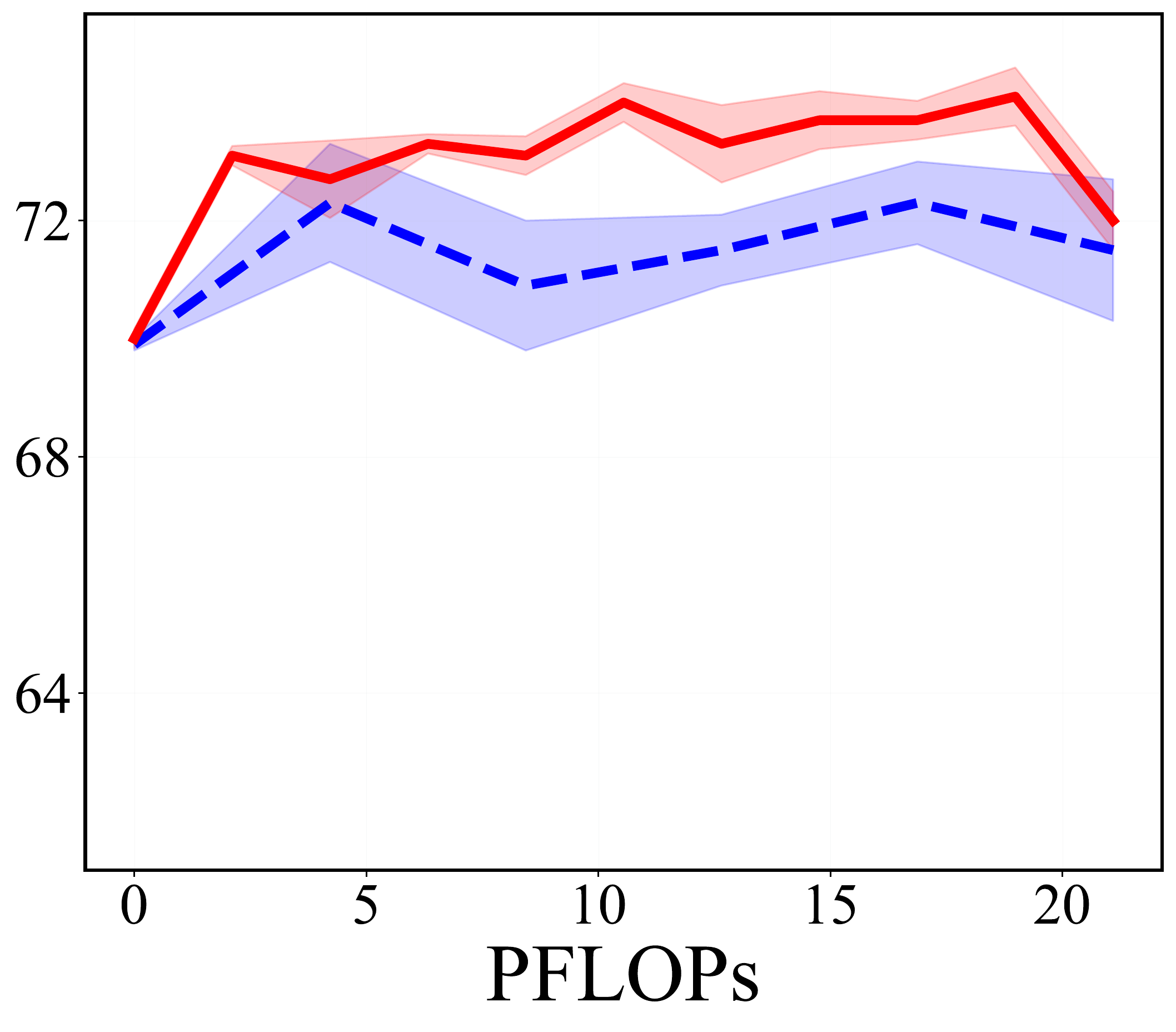}
    \caption{ImageNet-10 (Img/Cls=10)}
    \label{fig:imagenet10-10-flops}
  \end{subfigure}
  \caption{Performance comparison across varying training time and FLOPs.}
  \label{fig:efficiency-accuracy}
\end{figure*}

\subsection{Experimental Setups}

\noindent \textbf{Datasets.} We evaluate performance of neural networks trained on condensed datasets generated by several methods as baselines. Following previous works~\cite{DBLP:conf/iclr/ZhaoMB21,DBLP:conf/cvpr/Cazenavette00EZ22,DBLP:conf/icml/KimKOYSJ0S22}, we conduct experiments on both low- and high-resolution datasets including CIFAR-10, CIFAR-100, and ImageNet~\cite{DBLP:conf/cvpr/DengDSLL009}.

\noindent\textbf{Network\;Architectures.} Following previous works~\cite{DBLP:conf/icml/KimKOYSJ0S22,DBLP:journals/corr/abs-2110-04181}, we use a depth-3 ConvNet~\cite{DBLP:conf/iclr/SagunEGDB18} on CIFAR-10 and CIFAR-100. For ImageNet subsets, we follow IDC~\cite{DBLP:conf/icml/KimKOYSJ0S22} and adopt ResNetAP-10 for dataset distillation, a modified ResNet-10~\cite{DBLP:conf/cvpr/HeZRS16} by replacing strided convolution as average pooling for downsampling.

\noindent \textbf{Evaluation Metrics.} We study several methods in terms of performance and efficiency. The performance is measured by the testing accuracy of networks trained on condensed datasets. The efficiency is measured by GPU hours required by the dataset distillation process~\cite{DBLP:conf/aaai/FangMWSBZS22}. For a fair comparison, all GPU hours are measured on a single GPU. The training time of condensing CIFAR-10, CIFAR-100 and ImageNet subsets is evaluated on RTX-2080 GPU and RTX-A6000 GPU, respectively. We adopt FLOPs as a metric of computational efficiency.

\noindent \textbf{Baselines.} We compare our method with several prominent dataset condensation methods like (1) gradient-matching method including DSA~\cite{DBLP:conf/icml/ZhaoB21} and IDC~\cite{DBLP:conf/icml/KimKOYSJ0S22} (2) distribution-matching including DM~\cite{DBLP:journals/corr/abs-2110-04181} and CAFE~\cite{DBLP:conf/cvpr/WangZPZYWHBWY22} (3) parameter-matching including TM~\cite{DBLP:conf/cvpr/Cazenavette00EZ22b}. We use the state-of-the-art dataset distillation method IDC as the strongest baseline to calculate the gap between other methods on performance and efficiency.

\noindent\textbf{Training Details.} We adopt IDC as the backbone of our method, which is the state-of-the-art gradient-matching dataset distillation method. The outer loops and learning rate of condensed data are 400/100 and 0.01/0.1 for CIFAR-10/100 and ImageNet-Subsets. We employ 5/10 pre-trained models for CIFAR-10/100 and ImageNet. The number of pre-train epochs is 2/5/10 for CIFAR-10/100, ImageNet-10, and ImageNet-100. The setting of other hyperparameters follows IDC~\cite{DBLP:conf/icml/KimKOYSJ0S22} including the number of inner loops, batch size, and augmentation strategy.

\subsection{Condensed Data Evaluation}

\noindent\textbf{CIFAR-10\;\&\;CIFAR-100.} Our method achieves a better trade-off in task performance vs. the amount of training time and computation compared to other state-of-the-art baselines on CIFAR-10 and CIFAR-100. For instance, as shown in \cref{tab:cifar}, our method is comparable to 
IDC while achieving $\boldsymbol{5\times}$ and $\boldsymbol{10\times}$ \textbf{speed ups} on CIFAR-10. Our method shows $10\%$, $9\%$, and $7\%$ performance improvements over IDC on CIFAR-100 while achieving $\boldsymbol{5\times}$, $\boldsymbol{10\times}$, and $\boldsymbol{20\times}$ \textbf{acceleration}, respectively.

To further demonstrate the advantages of our method, we report the evaluation results across a varying amount of computational resources in the form of the number of training steps in \cref{fig:epoch-accuracy}, training time, and FLOPs in \cref{fig:efficiency-accuracy}. We observe that our method consistently outperforms all the baselines across different training steps, training times, and FLOPs. This demonstrates the effectiveness of our distillation method in capturing informative features from early-stage training; and enhanced diversity of the models for better generalizability. Interestingly, our method obtains better performance and efficiency over state-of-the-art baselines on CIFAR-100 as compared to CIFAR-10. This demonstrates the effectiveness and scalability of our method on large-scale datasets which makes it more appealing for all practical purposes.

\noindent\textbf{ImageNet.} Apart from CIFAR-10/100, we further investigate the performance and efficiency of our method on the high-resolution dataset ImageNet. Following previous baselines~\cite{DBLP:conf/eccv/TianKI20,DBLP:conf/icml/KimKOYSJ0S22}, we evaluate our method on ImageNet-subset consisting of 10 and 100 classes.

We observe that the dataset distillation methods on ImageNet suffer from severe efficiency challenges. As shown in \cref{tab:imagenet}, dataset distillation method IDC~\cite{DBLP:conf/icml/KimKOYSJ0S22} achieves high performance while requiring almost 4 days on ImageNet-10; while DSA~\cite{DBLP:conf/iclr/ZhaoMB21} and DM~\cite{DBLP:conf/icml/ZhaoB21} are more efficient in training time with significantly poor performance. The accuracy of networks trained on condensed data generated by our method outperforms all existing state-of-the-art baselines with the least training time. For instance, our method requires \textbf{less than 1 day} to condense ImageNet-10, which leads to $\boldsymbol{5\times}$ \textbf{speedup} over SOTA methods.

As shown in \cref{fig:epoch-accuracy} and \cref{fig:efficiency-accuracy}, we conduct extensive experiments with various training budgets. The results demonstrate that our method requires significantly fewer training steps, time, and computation resources to reach the same performance as the SOTA method IDC and achieves higher performance with the same training budgets. This indicates that utilizing early-stage models as initialization guides dataset distillation to focus on distinguishing features at the beginning of distillation. The exploration of diversity expands the parameter space and reduces the amount of time on learning repeated and redundant features.


\begin{table}[!ht]        
\vspace{2mm}
  \begin{subtable}[htbp]{1.\linewidth}
  \centering
  \resizebox*{8.3cm}{!}{
  \begin{tabular}{cc|ccc}
  \toprule
    \multirow{2}{*}{Dataset} & \multirow{2}{*}{Method} & \multicolumn{3}{c}{Evaluation model} \\
    {} & {} & {ConvNet-3} & {ResNet-10} & {DenseNet-121} \\
    \midrule
    \multirow{2}{*}{CIFAR-100} & {IDC~\cite{DBLP:conf/icml/KimKOYSJ0S22}}  & {45.1} & {\textbf{38.9}} & {39.5}\\
    {} & {$\text{Ours}_5$} & \cellcolor{mygray} {\textbf{46.5}} & \cellcolor{mygray} {\textbf{38.4}} & \cellcolor{mygray} {\textbf{39.6}} \\
    \bottomrule
  \end{tabular}
  }
  \caption{The performance of condensed CIFAR-100 dataset (10 images per class) trained on ConvNet-3 on different network architectures.}
  \end{subtable}
  \begin{subtable}[htbp]{1.\linewidth}
  \centering
  \resizebox*{8.3cm}{!}{
  \begin{tabular}{cc|ccc}
  \toprule
    \multirow{2}{*}{Dataset} & \multirow{2}{*}{Method} & \multicolumn{3}{c}{Evaluation model} \\
    {} & {} & {ResNetAP-10} & {ResNet-18}  & {EfficientNet-B0} \\
    \midrule
    \multirow{2}{*}{ImageNet-10} & {IDC~\cite{DBLP:conf/icml/KimKOYSJ0S22}}  & {74.0} & {73.1} & {74.3} \\
    {} & {$\text{Ours}_5$} & \cellcolor{mygray} {\textbf{74.6}} & \cellcolor{mygray} {\textbf{74.5}} & \cellcolor{mygray} {\textbf{75.4}}\\
    \bottomrule
  \end{tabular}
  }
  \caption{The performance of condensed ImageNet-10 dataset (10 images per class) trained on ResNetAP-10 on different network architectures.}
  \end{subtable}
  \caption{Performance of synthetic data learned on CIFAR100 and ImageNet-10 datasets with different architectures. The networks are trained on condensed dataset and validated on test dataset.}
  \label{tab:crossarchitecture}
\end{table}

\noindent \textbf{Cross-Architecture Generalization.} We also evaluate the performance of our condensed data on architectures different from the one used to distill it on the CIFAR-100 (1 and 10 images per class) and ImageNet-10 (10 images per class). In \cref{tab:crossarchitecture}, we show the performance of our baselines ConvNet-3 and ResNetAP-10 evaluated on ResNet-18~\cite{DBLP:conf/cvpr/HeZRS16}, DenseNet-121~\cite{DBLP:conf/cvpr/HuangLMW17}, and EfficientNet-B0~\cite{DBLP:conf/icml/TanL19}. 

For IDC~\cite{DBLP:conf/icml/KimKOYSJ0S22}, we use condensed data provided by the official implementation for evaluation of their method. Our method obtains the best performance on all the transfer models except for ResNet-10 on CIFAR-100 (10 images per class) where we lie within one standard deviation of IDC -- demonstrating the robustness of our method to changes in network architecture.

\subsection{Analysis}
\label{sec:analysis}

We perform ablation studies on our efficient dataset distillation method described in \cref{sec:method}. Specifically, we measure the impact of (1) the number of epochs of pre-training on real data, (2) the magnitude of parameter perturbation, (3) the number of early-stage models, and (4) the acceleration of training.

\begin{figure}[!ht]
  \centering
  \vspace{2mm}
    \begin{subfigure}{0.495\linewidth}
    \includegraphics[width=1.\linewidth]{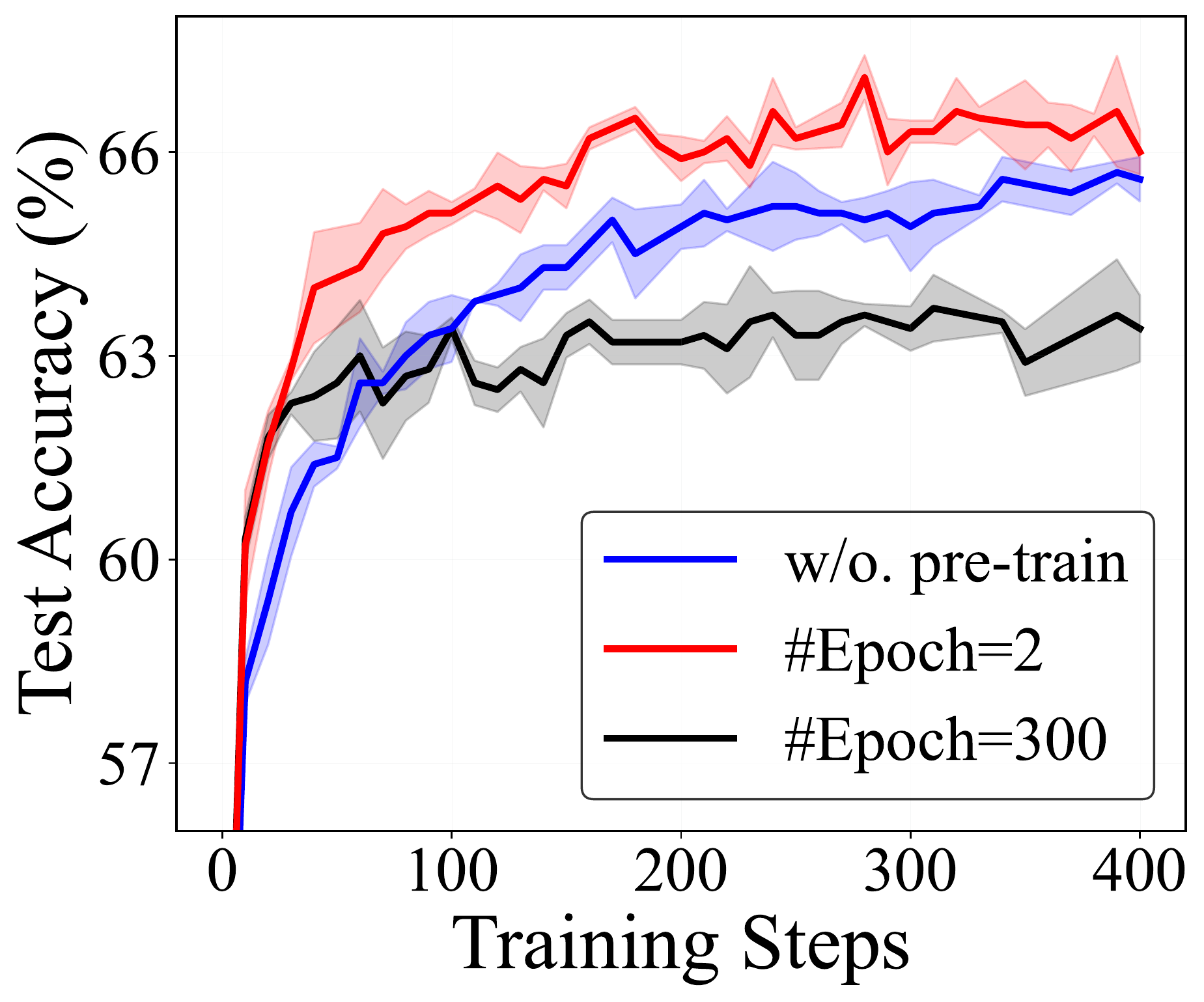}
    \caption{Effect of pre-train epochs}
    \label{fig:pretrain-ablation}
  \end{subfigure}
  \hfill
  \begin{subfigure}{0.495\linewidth}
    \includegraphics[width=1\linewidth]{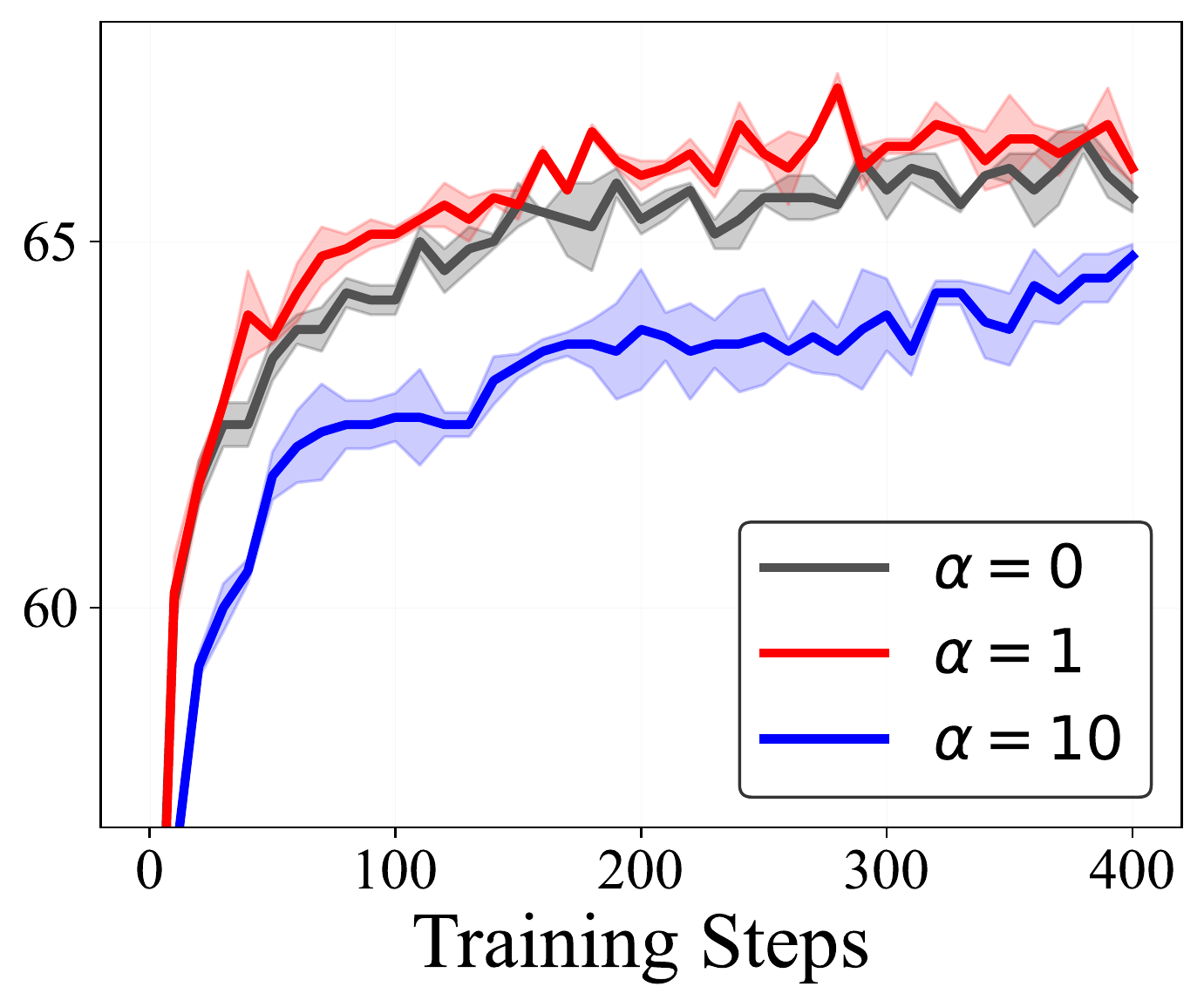}
    \caption{Effect of perturbation magnitude}
    \label{fig:magnitude-ablation}
  \end{subfigure}
   \caption{Condensation performance from networks pre-trained for different epochs and varying magnitudes of parameter perturbation. The networks are trained with same hyper-parameters except for training epochs and perturbation magnitudes, respectively. Evaluation is performed on CIFAR-10 (10 images per class).}
   \label{fig:magnitude-pretrain-ablation}
\end{figure}

\noindent \textbf{Epochs of Pre-training.} We study the effect of pre-training epochs on networks used in our method in terms of test accuracy on CIFAR-10 (10 images per class) and demonstrate results in \cref{fig:pretrain-ablation}. We observe that early-stage networks pre-trained with 2 epochs perform significantly better than randomly initialized networks and well-trained networks with 300 epochs. The results demonstrate that early-stage networks contain a more informative parameter space than randomly initialized networks, thereby helping the condensed datasets to capture features more efficiently. While it is generally known that well-trained networks perform better, well-trained networks tend to get stuck in local optima and lack diversity among parameter spaces. On the other hand, early-stage models provide flexible and informative guidance for dataset distillation.





\noindent \textbf{Magnitude of Parameter Perturbation.} We study the effect of the magnitude $\alpha$ of parameter perturbation in terms of test accuracy on CIFAR-10 (10 images per class) and report results in \cref{fig:magnitude-ablation}. We observe that condensed dataset achieves better performance on both accuracy and efficiency when magnitude $\alpha$ is carefully set as shown in \cref{fig:magnitude-ablation}. When the magnitude is large, \eg, 10, the perturbed networks diverge from the original space; the perturbed parameter space contains less relevant and inconsistent information, thereby impacting performance and efficiency. When the magnitude is small, such as not employing parameter perturbation, the parameter space lacks diversity compared to well-designed perturbed parameter space. Experimental results show that $\alpha=1$ is optimal for CIFAR in our setting which works consistently better across all training steps. Well-designed magnitude makes perturbed networks concentrated around the original network, thereby augmenting the parameter space with diversified and relevant information.

\begin{figure}[!ht]
  \centering
   \includegraphics[width=.8\linewidth]{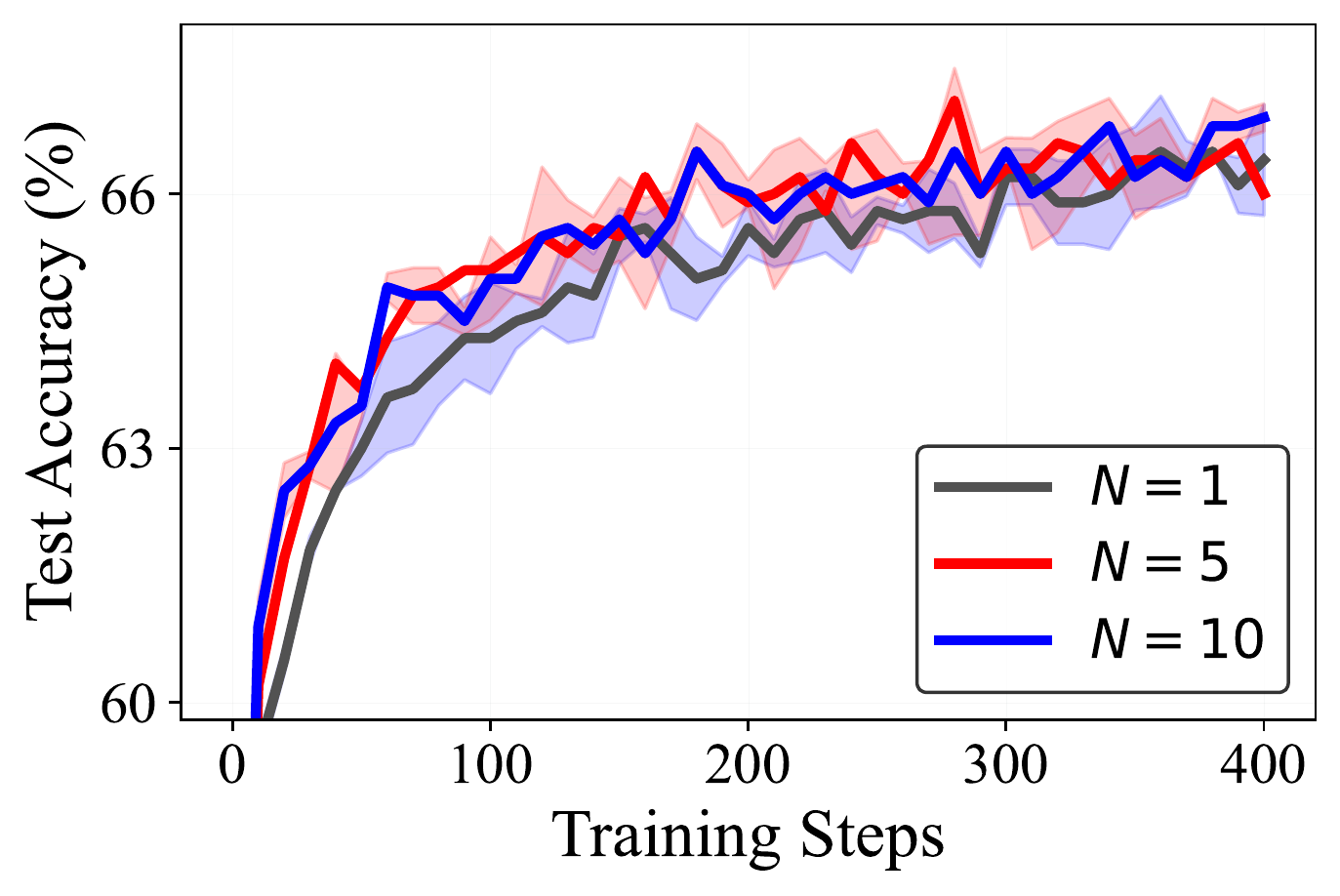}
   \caption{Condensation performance from a varying number of early-stage models. Performances with a varying number of models are similar, which demonstrates that our method is not sensitive to the number of models to achieve high performance.}
   \label{fig:num-ablation}
\end{figure}

\noindent \textbf{Number of Early-Stage Models.} We study the effect of the number of early-stage models in our experiment and show the results in \cref{fig:num-ablation}. It is observed that the number of early-stage models $N$ has less impact on the test accuracy of the condensed dataset. We argue that parameter perturbation in our method plays an important role in exploring the diversity of early-stage models; such that the description of parameter space depends on the representation of models rather than the number of models. In our method, a few models, \eg 5, can achieve comparable performance to SOTA~\cite{DBLP:conf/icml/KimKOYSJ0S22}, with two significant advantages. The first is to shorten training time as the number of outer loops in DD is closely related to the number of models $N$. The second is to reduce computation resources in network pre-training. TM~\cite{DBLP:conf/cvpr/Cazenavette00EZ22} also utilizes network pre-training in DD, however, the number of models in their method is relatively large, \eg 50, which is 10$\times$ more than ours. Parameter perturbation in our method augments the diversity of models and improves efficiency with only a small number of models.

\begin{figure}[!ht]
  \centering
   \includegraphics[width=.8\linewidth]{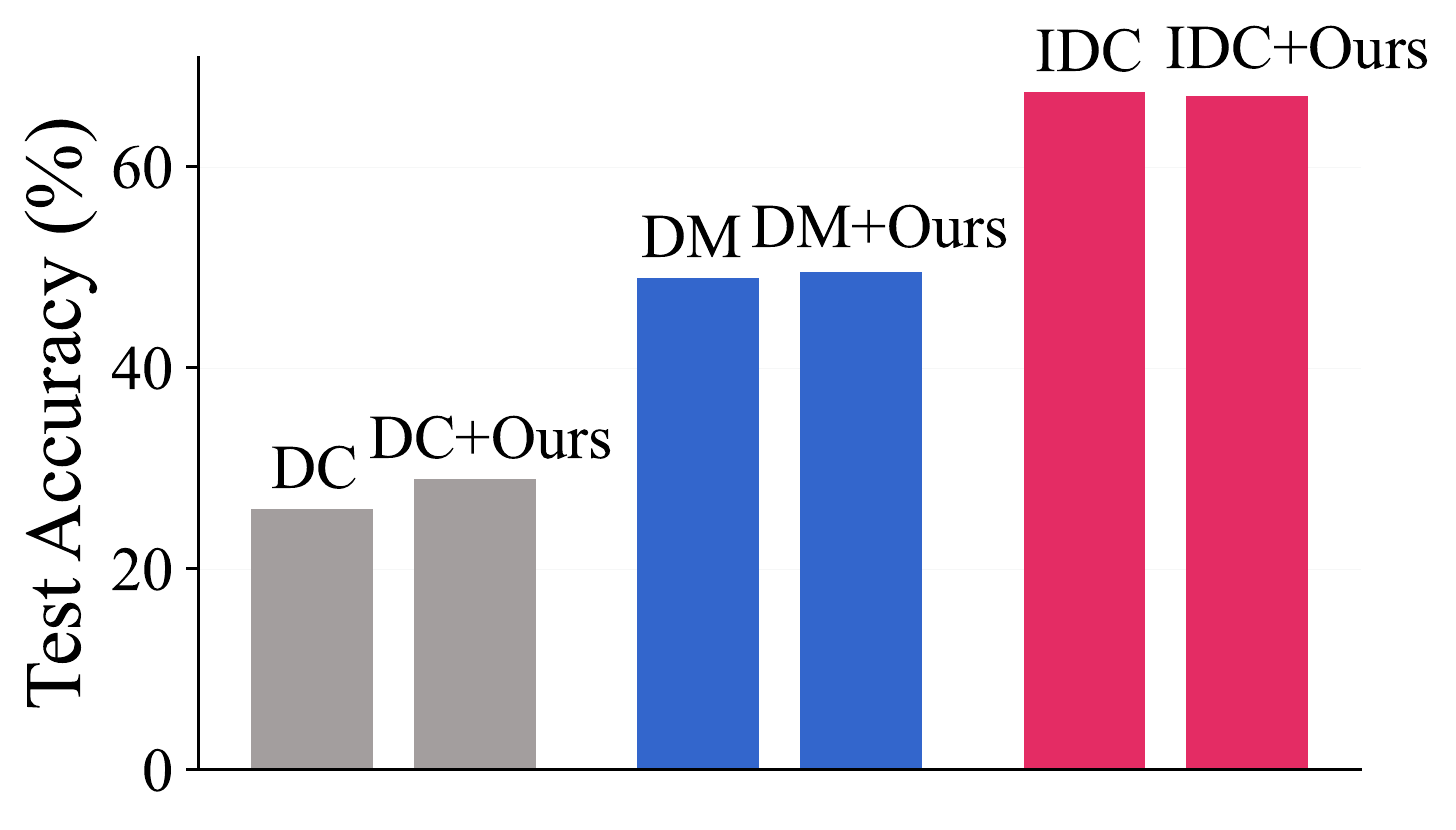}
   \caption{Performance of our method applied to different dataset distillation methods on CIFAR-10 dataset (10 images per class). Our results are reported with \textbf{$5\times$ training acceleration}.}
   \label{fig:acc-performance}
\end{figure}

\begin{table}[!ht]
    \centering
    \begin{subtable}[htbp]{1.\linewidth}
    \resizebox*{8.3cm}{!}{
    \begin{tabular}{c|cccc}
    \toprule
    {Speed up} & {DC~\cite{DBLP:conf/iclr/ZhaoMB21}} & {DSA~\cite{DBLP:conf/icml/ZhaoB21}} & {IDC~\cite{DBLP:conf/icml/KimKOYSJ0S22}} & {Ours}\\
    \midrule
    {$1\times$} & {44.9} & {52.1} & {67.5} & {-}\\
    {$5\times$} & {41.6 (-3.3)} & {47.0 (-5.1)} & {66.2 (-1.3)} & {67.1}\\
    {$10\times$} & {39.2 (-5.7)} & {46.2 (-5.9)} & {65.0 (-2.5)} & {66.5 (\textbf{-0.6})}\\
    {$20\times$} & {37.8 (-7.1)} & {44.8 (-7.3)} & {63.7 (-3.8)} & {65.2 (\textbf{-1.9})}\\
    \bottomrule
    \end{tabular}
    }
    \caption{CIFAR-10 (Img/Cls=10)}
    \end{subtable}
    
    \begin{subtable}[htbp]{1.\linewidth}
    \resizebox*{8.3cm}{!}{
    \begin{tabular}{c|cccc}
    \toprule
    {Speed up} & {DC~\cite{DBLP:conf/iclr/ZhaoMB21}} & {DSA~\cite{DBLP:conf/icml/ZhaoB21}} & {IDC~\cite{DBLP:conf/icml/KimKOYSJ0S22}} & {Ours}\\
    \midrule
    {$1\times$} & {53.9} & {60.6} & {74.5} & {-}\\
    {$5\times$} & {50.3 (-3.6)} & {56.5 (-4.1)} & {73.3 (-1.2)} & {73.8}\\
    {$10\times$} & {47.3 (-6.6)} & {55.7 (-4.9)} & {72.0 (-2.5)} & {73.1 (\textbf{-0.7})}\\
    {$20\times$} & {42.0 (-11.9)} & {54.1 (-6.5)} & {71.1 (-3.4)} & {71.7 (\textbf{-2.1})}\\
    \bottomrule
    \end{tabular}
    }
    \caption{CIFAR-10 (Img/Cls=50)}
    \end{subtable}


    \begin{subtable}[htbp]{1.\linewidth}
    \resizebox*{8.3cm}{!}{
    \begin{tabular}{c|cccc}
    \toprule
    {Speed up} & {DC~\cite{DBLP:conf/iclr/ZhaoMB21}} & {DSA~\cite{DBLP:conf/icml/ZhaoB21}} & {IDC~\cite{DBLP:conf/icml/KimKOYSJ0S22}} & {Ours}\\
    \midrule
    {$1\times$} & {29.5} & {32.3} & {45.1} & {-}\\
    {$5\times$} & {23.1 (-6.4)} & {29.3 (-3.0)} & {43.4 (-1.9)} & {46.2}\\
    {$10\times$} & {21.1 (-8.4)} & {28.7 (-3.6)} & {41.6 (-3.5)} & {45.6 (\textbf{-0.6})}\\
    {$20\times$} & {18.6 (-10.9)} & {27.9 (-4.4)} & {40.5 (-4.6)} & {45.0 (\textbf{-1.2})}\\
    \bottomrule
    \end{tabular}
    }
    \caption{CIFAR-100 (Img/Cls=10)}
    \end{subtable}
    
    \caption{Condensation performance with different acceleration / speed ups over state-of-the-art dataset distillation approaches. We show performance drop between increased speed up in brackets. Our method achieves higher performance over baseline methods at all levels of speed up. With increased speed up, our method shows minor regression in performance.}
    \label{tab:acceleration-acc}
    \vspace{-2mm}
\end{table}

\noindent \textbf{Acceleration of Training.} We study the effect of acceleration of training on existing DD methods~\cite{DBLP:conf/icml/ZhaoB21,DBLP:conf/iclr/ZhaoMB21,DBLP:conf/icml/KimKOYSJ0S22} and our method. We observe our method to retain similar performance with minor regression to increased training acceleration / speed-ups -- while the performance of existing methods drops dramatically in \cref{tab:acceleration-acc}. Our method achieves better performance than baselines at all levels of speed-up. This demonstrates the informativeness of our parameter space in terms of diversity and reduced redundancy; such that the condensed dataset does not learn similar information repeatedly and captures sufficient features efficiently. It is worth noting that our method performs better with less regression at higher levels of speed up on the more complex dataset, \eg, CIFAR-100. We also demonstrate our method can be orthogonally applied to other dataset distillation methods in \cref{fig:acc-performance}. We apply parameter perturbation on other DD methods to accelerate the training $5\times$ faster. This indicates better scalability and improved efficiency of our method in condensing large-scale datasets.

\section{Conclusion}
\label{sec:conclusion}

In this work, we introduce a novel method for improving the efficiency of gradient-matching based dataset distillation approaches. We leverage model augmentation strategies with early-stage training and parameter perturbation to increase the diversity of the parameter space as well as massively reduce the computation resource for dataset distillation. 
Our method is able to achieve $10\times$ acceleration on CIFAR and $5\times$ acceleration on ImageNet. As the first attempt to improve the efficiency of gradient-matching based dataset distillation, the proposed method successfully crafts a condensed dataset of ImageNet in 18 hours, making dataset distillation more applicable in real-world settings. 


{\small
\bibliographystyle{ieee_fullname}
\bibliography{egbib}
}
\clearpage

\end{document}


\title{Accelerating Dataset Distillation via Model Augmentation}

\author{
Lei Zhang\textsuperscript{1*} \hspace{2mm}
Jie Zhang\textsuperscript{1*} \hspace{2mm}
Bowen Lei\textsuperscript{2} \hspace{2mm} 
Subhabrata Mukherjee\textsuperscript{3} \hspace{2mm} \\
Xiang Pan\textsuperscript{4} \hspace{2mm}
Bo Zhao\textsuperscript{5} \hspace{2mm}
Caiwen Ding\textsuperscript{6} \hspace{2mm}
Yao Li\textsuperscript{7} \hspace{2mm}
Dongkuan Xu\textsuperscript{8$\dagger$} \\
\textsuperscript{1}Zhejiang University \hspace{3mm} \textsuperscript{2}Texas A\&M University \hspace{3mm} \textsuperscript{3}Microsoft Research \\ \textsuperscript{4}New York University \hspace{2mm} \textsuperscript{5}Beijing Academy of Artificial Intelligence \hspace{2mm} \textsuperscript{6}University of Connecticut \\ \textsuperscript{7}University of North Carolina, Chapel Hill \hspace{3mm} \textsuperscript{8}North Carolina State University  \\
{\tt\small \{zl\_leizhang, zj\_zhangjie\}@zju.edu.cn} \hspace{3mm}\tt\small dxu27@ncsu.edu
}

\newcommand{\customfootnotetext}[2]{{
\renewcommand{\thefootnote}{#1}
\footnotetext[0]{#2}}}

\maketitle

This supplementary material provides more details on the method and experiments, including a detailed explanation of early-stage models from gradient perspective (\cref{sec:earlystage}), datasets (\cref{sec:datasets}), network architectures (\cref{sec:networks}), and additional experiment results (\cref{sec:experiments}).

\begin{table}[!ht]
    \vspace{-2mm}
    \centering
    \begin{tabular}{ccc}
    \toprule
    {Parameter} & {Shape} & {Layer hyper-parameter}\\
    \midrule
    {pooling.avg} & {[2]} & {stride=2, padding=0}\\
    {conv1.weight} & {[3, 128, 3, 3]} & {stride=1, padding=1}\\
    {conv1.bias} & {[128]} & {N/A}\\
    {conv2.weight} & {[128, 128, 3, 3]} & {stride=1, padding=1}\\
    {conv2.bias} & {[128]} & {N/A} \\
    {conv2.weight} & {[128, 128, 3, 3]} & {stride=1, padding=1}\\
    {conv2.bias} & {[128]} & {N/A}\\
    {norm.group} & {[128, 128]} & {eps=1e-5, affine=True}\\
    {norm.group} & {[128, 128]} & {eps=1e-5, affine=True}\\
    {norm.group} & {[128, 128]} & {eps=1e-5, affine=True}\\
    {pooling.avg} & {[2]} & {stride=2, padding=0}\\
    {pooling.avg} & {[2]} & {stride=2, padding=0}\\
    {pooling.avg} & {[2]} & {stride=2, padding=0}\\
    {fc.weight} & {[2048, 10]} & {N/A}\\
    {fc.bias} & {[10]} & {N/A}\\
    \bottomrule
    \end{tabular}
    \caption{Detailed information of the ConvNet-3 architecture used in our experiments for CIFAR-10 and CIFAR-100.}
    \label{tab:convnet}
    \vspace{-5mm}
\end{table}

\section{Why Early-stage Models Work Better}
\label{sec:earlystage}
From the gradient perspective, early-stage models can produce diverse and large-magnitude gradients which are more effective for gradient matching. Recent studies~\cite{DBLP:conf/iclr/FrankleSM20,DBLP:journals/corr/abs-1812-04754} demonstrate that gradient dynamically converges to a very small subspace after a short period of training. The models and synthetic data will be alternatively updated after sampling the model in the dataset distillation process. Thus, successive gradients will be used for updating synthetic data. Training with small and similar successive gradients produced by well-trained models will result in worse synthetic data. This is consistent with the finding in the previous work, DSA~\cite{DBLP:conf/icml/ZhaoB21}, in which utilizes data augmentation to enlarge the gradient magnitude for better gradient matching.

\begin{table}[ht]
    \vspace{-2mm}
    \centering
    \begin{tabular}{ccc}
    \toprule
    {Parameter} & {Shape} & {Layer hyper-parameter} \\
    \midrule
    {conv1.weight} & {[3, 64, 7, 7]} & {stride=1, padding=3}\\
    {pool1.avg} & {[4, 4]} & {stride=4, padding=0} \\
    {conv2.weight} & {[64, 64, 3, 3]} & {stride=1, padding=1} \\
    {norm1.group} & {[64, 64]} & {eps=1e-5, affine=True}  \\
    {conv2.weight} & {[64, 64, 3, 3]} & {stride=1, padding=1} \\
    {norm1.group} & {[64, 64]} & {eps=1e-5, affine=True}  \\
    {conv3.weight} & {[64, 128, 3, 3]} & {stride=1, padding=1}\\
    {norm2.group} & {[128, 128]} & {eps=1e-5, affine=True} \\
    {conv4.weight} & {[128, 128, 3, 3]} & {stride=1, padding=1}\\
    {norm2.group} & {[128, 128]} & {eps=1e-5, affine=True} \\
    {conv5.weight} & {[64, 128, 1, 1]} & {stride=1, padding=0}\\
    {pool2.avg} & {[2, 2]} & {stride=2, padding=0} \\
    {norm2.group} & {[128, 128]} & {eps=1e-5, affine=True} \\
    {conv6.weight} & {[128, 256, 3, 3]} & {stride=1, padding=1} \\
    {norm3.group} & {[256, 256]} & {eps=1e-5, affine=True}  \\
    {conv7.weight} & {[256, 256, 1, 1]} & {stride=1, padding=1} \\
    {norm3.group} & {[256, 256]} & {eps=1e-5, affine=True}  \\
    {conv8.weight} & {[128, 256, 1, 1]} & {stride=1, padding=0} \\
    {pool2.avg} & {[2, 2]} & {stride=2, padding=0} \\
    {norm3.group} & {[256, 256]} & {eps=1e-5, affine=True}  \\
    {conv9.weight} & {[256, 512, 3, 3]} & {stride=1, padding=1} \\
    {norm4.group} & {[512, 512]} & {eps=1e-5, affine=True} \\
    {conv10.weight} & {[512, 512, 3, 3]} & {stride=1, padding=1} \\
    {norm4.group} & {[512, 512]} & {eps=1e-5, affine=True} \\
    {conv11.weight} & {[256, 512, 1, 1]} & {stride=1, padding=1} \\
    {pool2.avg} & {[2, 2]} & {stride=2, padding=0} \\
    {norm4.group} & {[512, 512]} & {eps=1e-5, affine=True} \\
    {pool5.avg} & {[7, 7]} & {stride=7, padding=0} \\
    {fc.weight} & {[512, 10]} & {N/A} \\
    {fc.bias} & {[10]} & {N/A} \\
    \bottomrule
    \end{tabular}
    \caption{Detailed information of the ResNetAP-10 architecture used in our experiments for ImageNet.}
    \label{tab:resnetap}
    \vspace{-5mm}
\end{table}

\vspace{-5mm}

\begin{table*}[!ht]
    \centering
    \begin{subtable}[htbp]{1.\linewidth}
    \centering
    \begin{tabular}{c|ccccccccccc}
    \toprule
    \multirow{2}{*}{\makecell{Img/\\Cls}} & \multicolumn{11}{c}{CIFAR-10} \\
    {} & {DM} & {DSA} & {CAFE} & {TM} &{$\text{IDC}_{\text{1}}$} & {$\text{IDC}_{\text{5}}$} & {$\text{Ours}_{\text{5}}$} & {$\text{IDC}_{\text{10}}$} & {$\text{Ours}_{\text{10}}$} & {$\text{IDC}_{\text{20}}$} & {$\text{Ours}_{\text{20}}$} \\
    \midrule
    {1} & {26.0} & {28.2} & {30.3} & {46.3} & {50.6} & {49.5 (0.4)} & {\textbf{49.2 (0.4)}} & {49.0 (0.3)} & {\textbf{48.7 (0.5)}} & {48.6 (0.3)} & {\textbf{48.0 (0.3)}}\\
    {10} & {48.9} & {52.1} & {46.3} & {65.3} & {67.5} & {66.2 (0.3)} & {\textbf{67.1 (0.2)}} & {65.0 (0.3)} & {\textbf{66.5 (0.1)}} & {63.7 (0.2)} & {\textbf{65.1 (0.1)}}\\ 
    {50} & {63.0} & {60.6} & {55.5} & {71.6} & {74.5} & {73.3 (0.3)} & {\textbf{73.8 (0.1)}} & {72.0 (0.6)} & {\textbf{73.1 (0.1)}} & {71.1 (0.2)} & {\textbf{71.7 (0.3)}}\\
    \bottomrule
    \end{tabular}
    \caption{Results on CIFAR-10}
    \end{subtable}
    \begin{subtable}[htbp]{1.\linewidth}
    \centering
    \begin{tabular}{c|ccccccccccc}
    \toprule
    \multirow{2}{*}{\makecell{Img/\\Cls}} & \multicolumn{11}{c}{CIFAR-100} \\
    {} & {DM} & {DSA} &  {CAFE} & {TM} & {$\text{IDC}_{\text{1}}$} & {$\text{IDC}_{\text{5}}$} & {$\text{Ours}_{\text{5}}$} & {$\text{IDC}_{\text{10}}$} & {$\text{Ours}_{\text{10}}$} & {$\text{IDC}_{\text{20}}$} & {$\text{Ours}_{\text{20}}$} \\
    \midrule
    {1} & {11.4} & {13.9} & {12.9} & {24.3} & {25.1} & {24.9 (0.5)} & {\textbf{29.8 (0.2)}} & {24.7 (0.1)} & {\textbf{29.6 (0.1)}} & {24.4 (0.1)} & {\textbf{29.1 (0.1)}}\\
    {10} & {29.7} & {32.3} & {27.8} & {40.1} & {45.1} & {44.1 (0.2)} & {\textbf{46.2 (0.3)}} & {43.1 (0.2)} & {\textbf{45.6 (0.4)}} & {41.6 (0.2)} & {\textbf{45.0 (0.0)}}\\ 
    {50} & {43.6} & {42.8} & {37.9} & {47.7} & {-} & {-} & {\textbf{52.6 (0.4)}} & {-} & {\textbf{52.3 (0.2)}} & {-} & {\textbf{52.2 (0.1)}}\\
    \bottomrule
    \end{tabular}
    \caption{Results on CIFAR-100}
    \end{subtable}
    \caption{Comparing performance of dataset distillation methods on CIFAR-10 and CIFAR-100. We report Top-1 test accuracy of test models ConvNet-3 trained on condensed datasets. Img/Cls means the number of images per class of the condensed dataset. We evaluate each task with 3 repetitions and denote the standard deviations in the parenthesis. We compare the same acceleration levels between IDC and our method on $5\times$, $10\times$, and $20\times$.}
    \label{tab:cifar-add}
\end{table*}

\begin{table*}[!ht]
    \centering
    \begin{tabular}{cc|ccccccc}
    \toprule
    {Dataset} & {Img/Cls} & {DM} & {DSA} & {$\text{IDC}_{\text{1}}$} & {$\text{IDC}_{\text{5}}$} & {$\text{Ours}_{\text{5}}$} & {$\text{IDC}_{\text{10}}$} & {$\text{Ours}_{\text{10}}$} \\ 
    \midrule
    \multirow{2}{*}{ImageNet-10} & {10} & {52.3} & {52.7} & {72.8 (0.6)} & {72.3 (0.7)} & {\textbf{74.6 (0.4)}} & {72.3 (1.0)} & {\textbf{74.0 (0.4)}} \\
    {} & {20} & {59.3} & {57.4} & {76.6 (0.4)} & {74.7 (0.4)} & {\textbf{76.3 (1.0)}} & {74.7 (0.4)} & {\textbf{75.2 (1.0)}} \\ 
    \midrule
    \multirow{2}{*}{ImageNet-100} & {10} & {22.3} & {21.8} & {46.7 (0.2)} & {-} & {\textbf{48.4 (0.3)}} & {-} & {-} \\
    {} & {20} & {30.4} & {30.7} & {53.7 (0.9)} & {-} & {\textbf{56.0 (0.5)}} & {-} & {-} \\ 
    \bottomrule
    \end{tabular}    
    \caption{Comparing performance of dataset distillation methods on ImageNet-10 and ImageNet-100. We report Top-1 test accuracy of test models ResNetAP-10 trained on condensed datasets. We evaluate each task with 3 repetitions and denote the standard deviations in the parenthesis.}
    \label{tab:imagenet-add}
\end{table*}

\section{Datasets}
\label{sec:datasets}

\noindent \textbf{ImageNet-subset.} Following previous works~\cite{DBLP:conf/eccv/TianKI20,DBLP:conf/icml/KimKOYSJ0S22}, we evaluate our method on ImageNet-subset, which borrows a subclass list containing 100 classes from ~\cite{DBLP:conf/eccv/TianKI20}. We use the first 10 classes from the list in our ImageNet-10 experiments and the complete list in our ImageNet-100 experiment. The images in ImageNet-subset are preprocessed to a fixed size of 224 $\times$ 224 using resize and center crop functions.

\section{Networks}
\label{sec:networks}

Staying with precdent~\cite{DBLP:conf/icml/ZhaoB21,DBLP:conf/iclr/ZhaoMB21,DBLP:conf/icml/KimKOYSJ0S22,DBLP:conf/cvpr/Cazenavette00EZ22}, we employ a simple ConvNet-3 architecture for CIFAR-10 and CIFAR-100~\cite{cifar10} dataset and a modified ResNet-10~\cite{DBLP:conf/cvpr/HeZRS16} architecture ResNetAP-10 for ImageNet-subset. As shown in ~\cref{tab:convnet}, ConvNet-3 consists of several convolutional tasks, each containing a 3 $\times$ 3 convolutional layer with 129 filters, Instance normalization, RELU and 2 $\times$2 average pooling with stride 2. We also demonstrate the detailed architecture of ResNetAP-10 in ~\cref{tab:resnetap}. ResNetAP-10 is modified on ResNet-10, which replaces strided convolution as average pooling for downsampling.

\section{Additional Experiments}
\label{sec:experiments}

\subsection{More Experiment results on Datasets}

\noindent \textbf{CIFAR-10 \& CIFAR-100.} Our method achieves a better trade-off in task performance vs. acceleration of training compared to other state-of-the-art baselines on CIFAR-10 and CIFAR-100. As shown in \cref{tab:cifar-add}, our method consistently outperforms SOTA method IDC under $5\times$, $10\times$, and $20\times$ speed-ups and other baselines without acceleration. This verifies the efficiency of our method, which requires less training time and computation resources to achieve comparable or surpass performances of leading baselines. 

\begin{figure*}[ht]
  \centering
  \begin{subfigure}{0.235\linewidth}
    \includegraphics[width=1\linewidth]{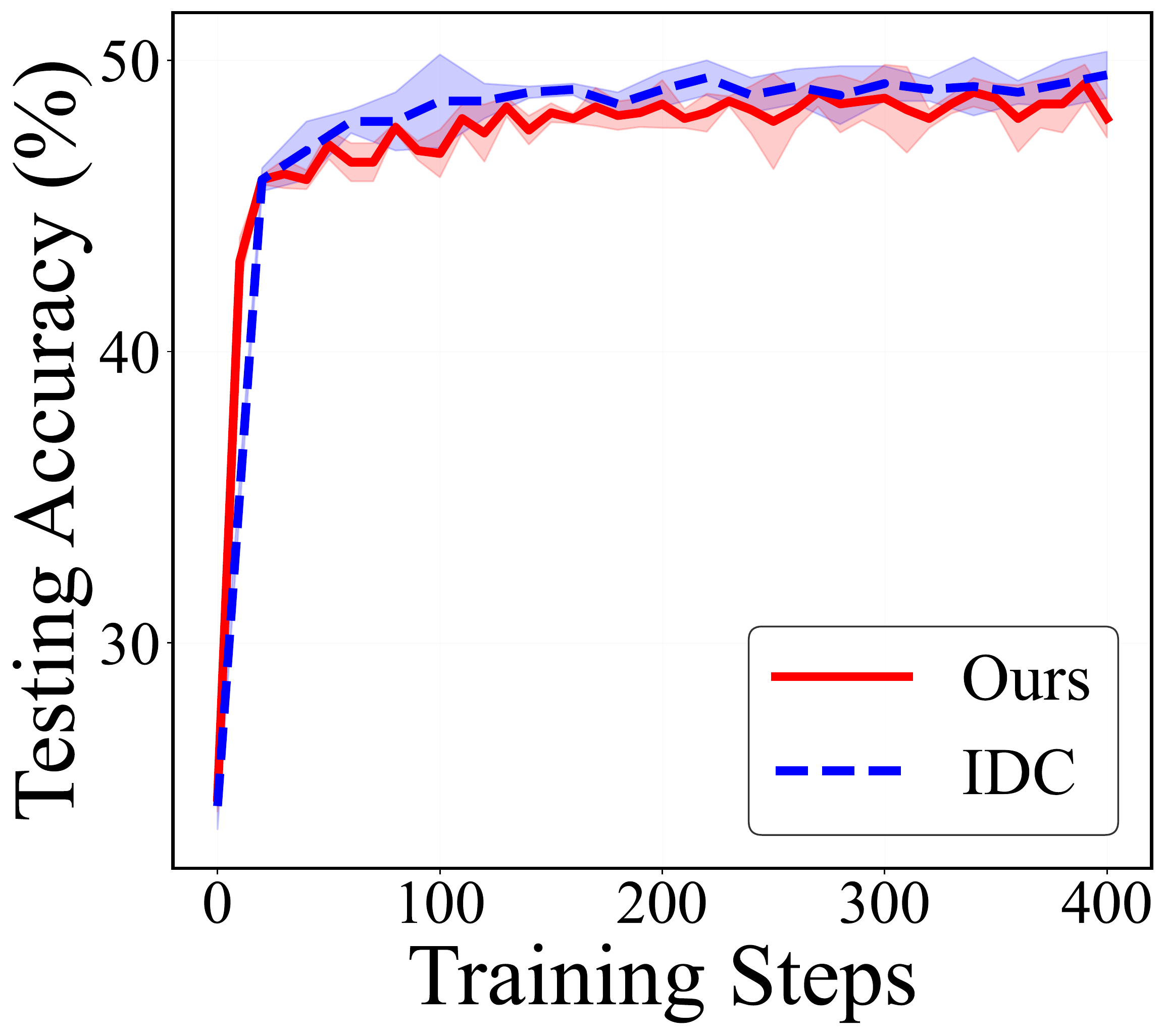}
    \caption{CIFAR-10 (Img/Cls=1).}
    \label{fig:cifar10-1-epoch}
  \end{subfigure}
  \hfill
  \begin{subfigure}{0.235\linewidth}
    \includegraphics[width=1.\linewidth]{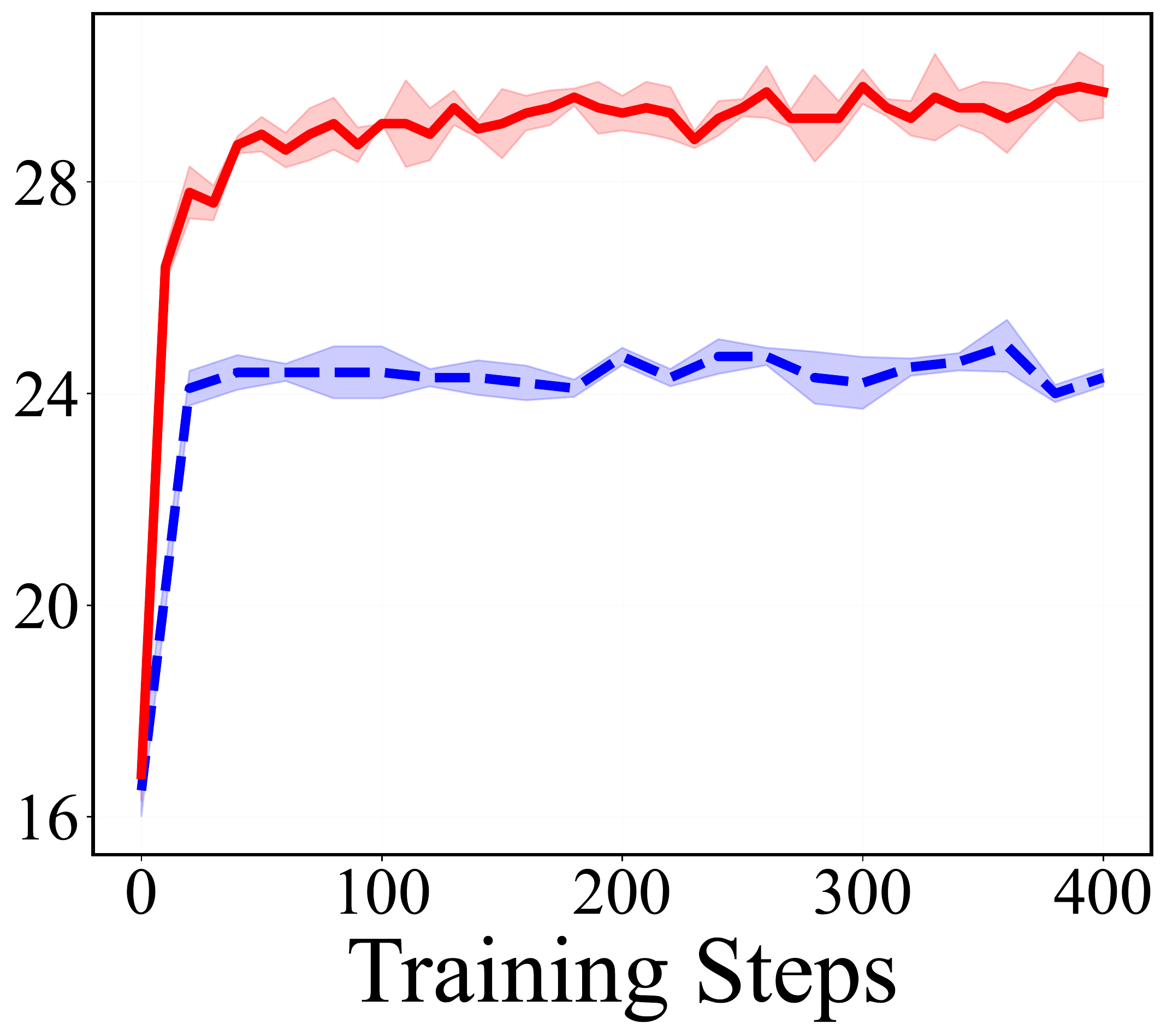}
    \caption{CIFAR-100 (Img/Cls=1)}
    \label{fig:cifar100-1-epoch}
  \end{subfigure}
  \hfill
  \begin{subfigure}{0.235\linewidth}
    \includegraphics[width=1.\linewidth]{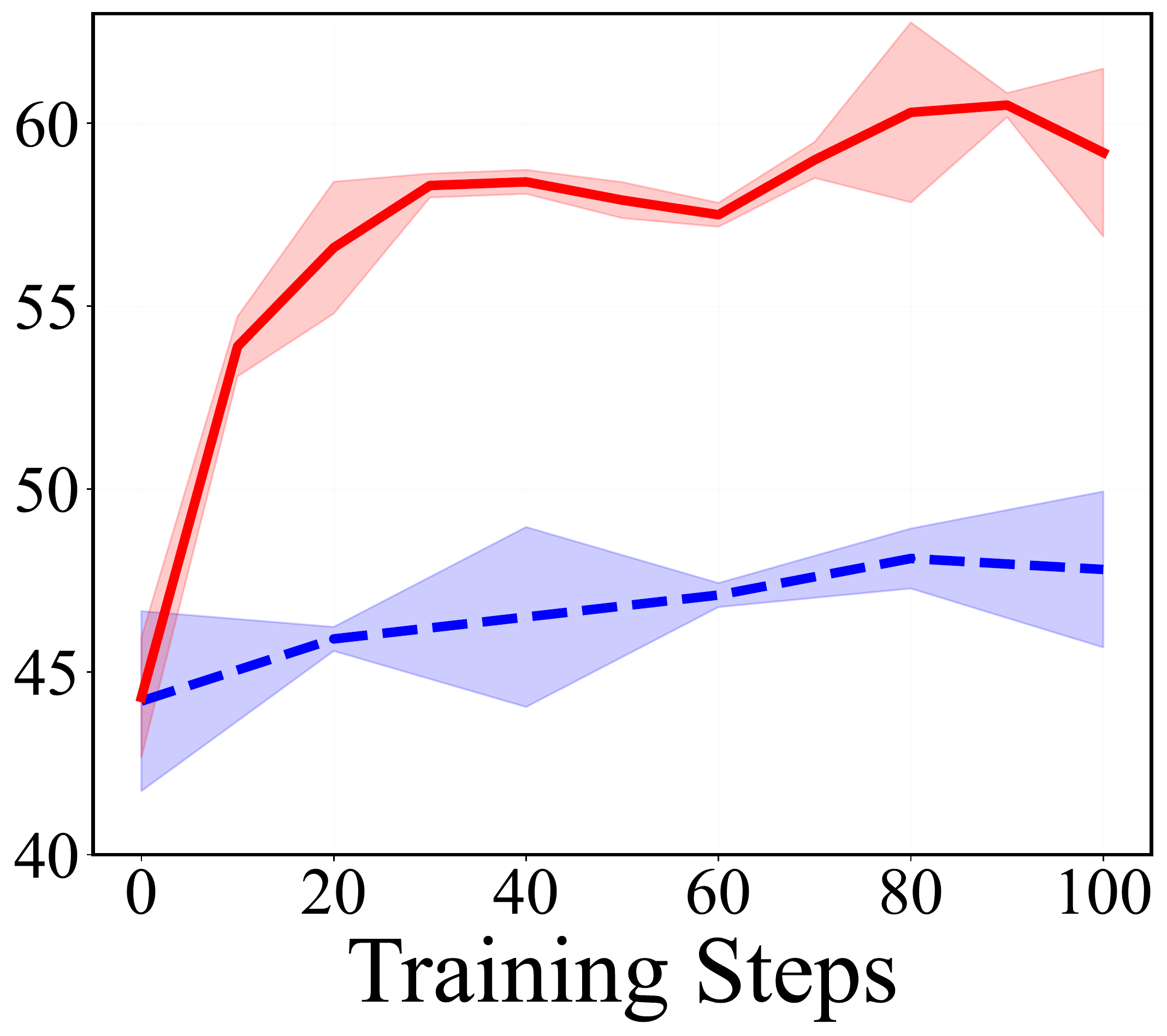}
    \caption{ImageNet-10 (Img/Cls=1)}
    \label{fig:imagenet10-1-epoch}
  \end{subfigure}
  \hfill
  \begin{subfigure}{0.235\linewidth}
    \includegraphics[width=1.\linewidth]{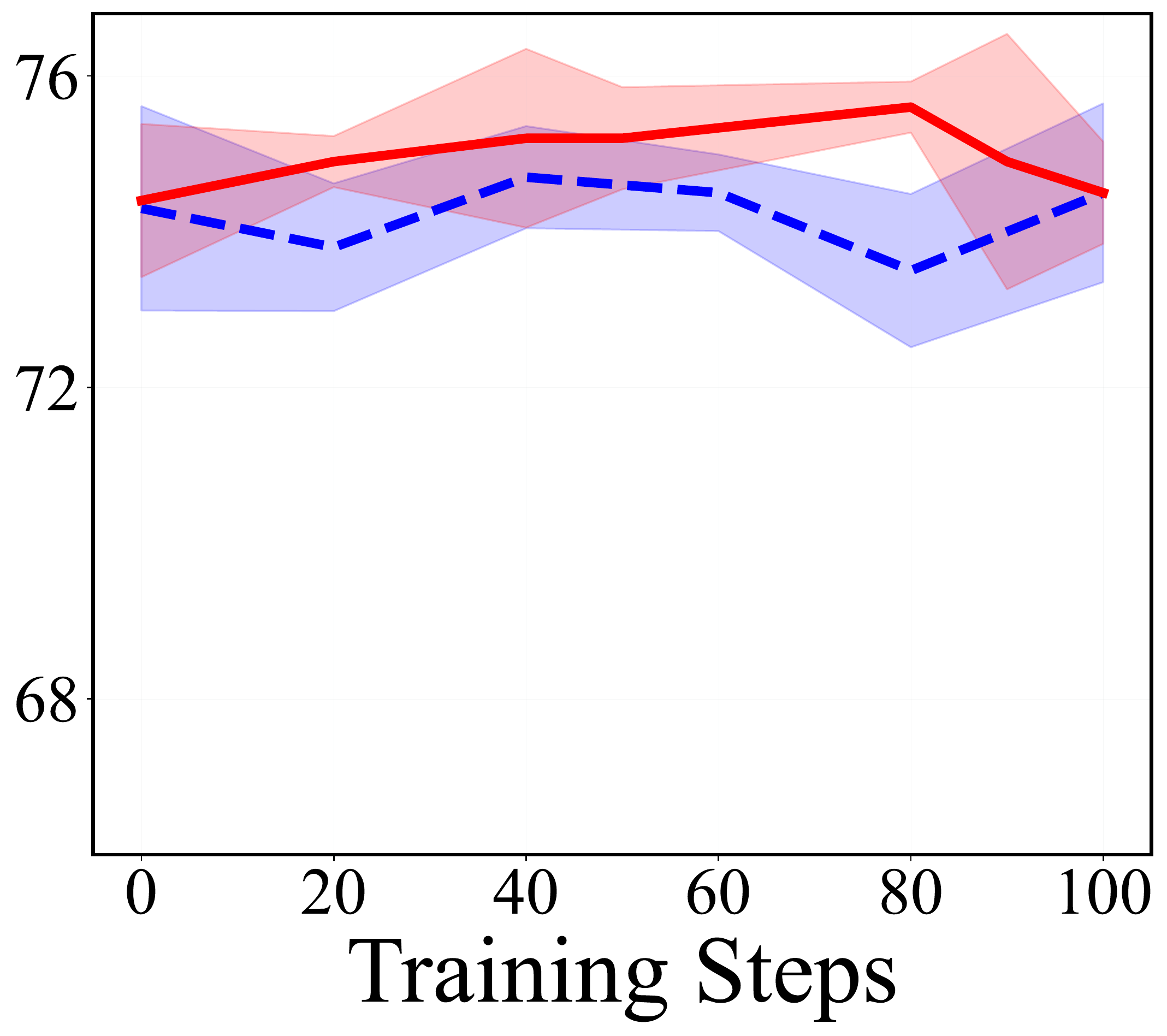}
    \caption{ImageNet-10 (Img/Cls=20)}
    \label{fig:imagenet10-20-epoch}
  \end{subfigure}
  \caption{Performance comparison across varying training steps. The batch size is set the same as 64. Our method consistently outperforms the leading method IDC~\cite{DBLP:conf/icml/KimKOYSJ0S22} on various training steps.}
  \label{fig:epoch-accuracy-add}
  \vspace{-3mm}
\end{figure*}

\begin{figure*}[ht]
  \centering
  \begin{subfigure}{0.235\linewidth}
    \includegraphics[width=1\linewidth]{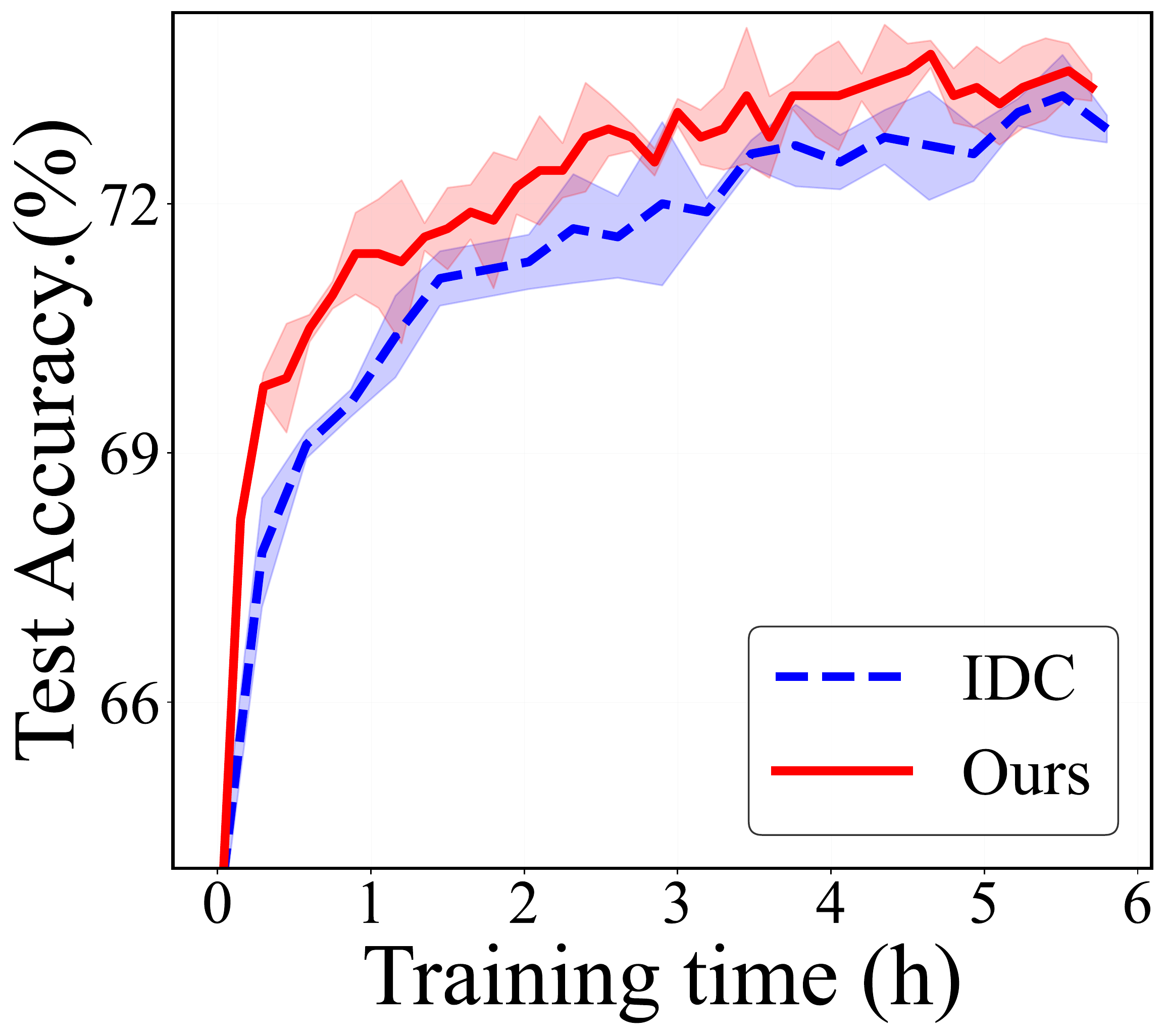}
    \caption{CIFAR-10 (Img/Cls=50).}
    \label{fig:cifar10-50-time}
  \end{subfigure}
  \hfill
  \begin{subfigure}{0.235\linewidth}
    \includegraphics[width=1.\linewidth]{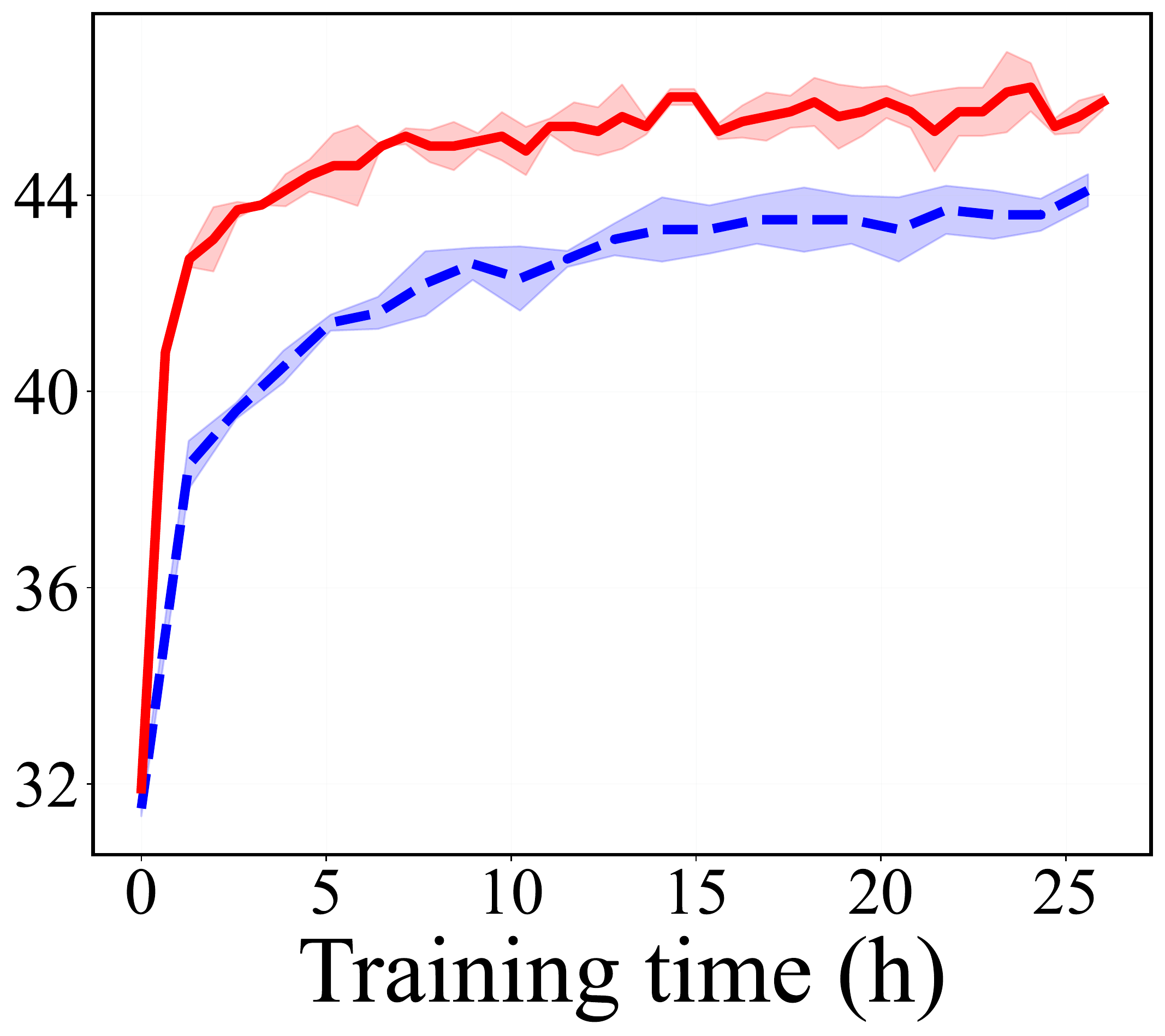}
    \caption{CIFAR-100 (Img/Cls=10)}
    \label{fig:cifar100-10-time}
  \end{subfigure}
  \hfill
  \begin{subfigure}{0.235\linewidth}
    \includegraphics[width=1.\linewidth]{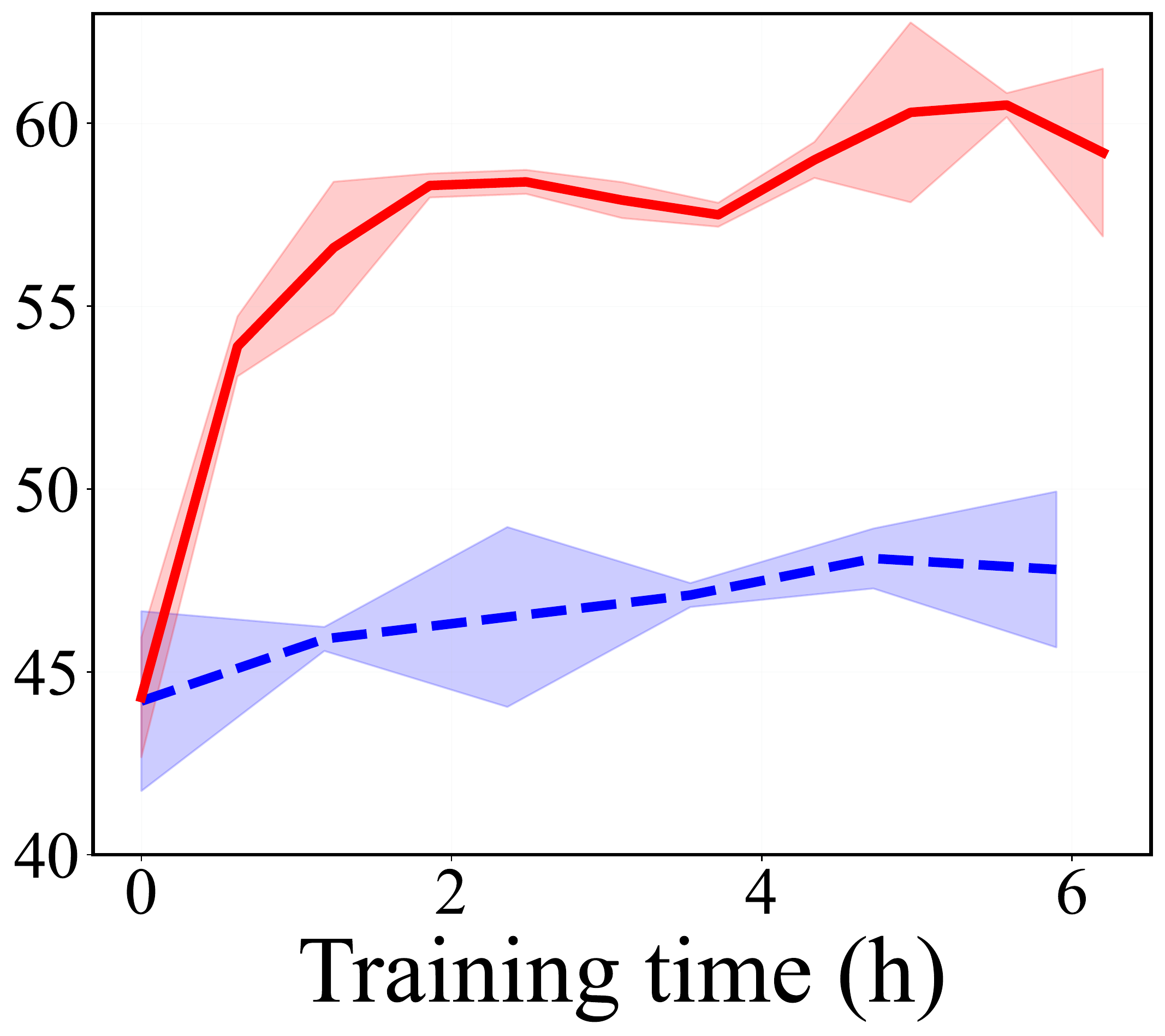}
    \caption{ImageNet-10 (Img/Cls=1)}
    \label{fig:imagenet10-1-time}
  \end{subfigure}
  \hfill
  \begin{subfigure}{0.235\linewidth}
    \includegraphics[width=1.\linewidth]{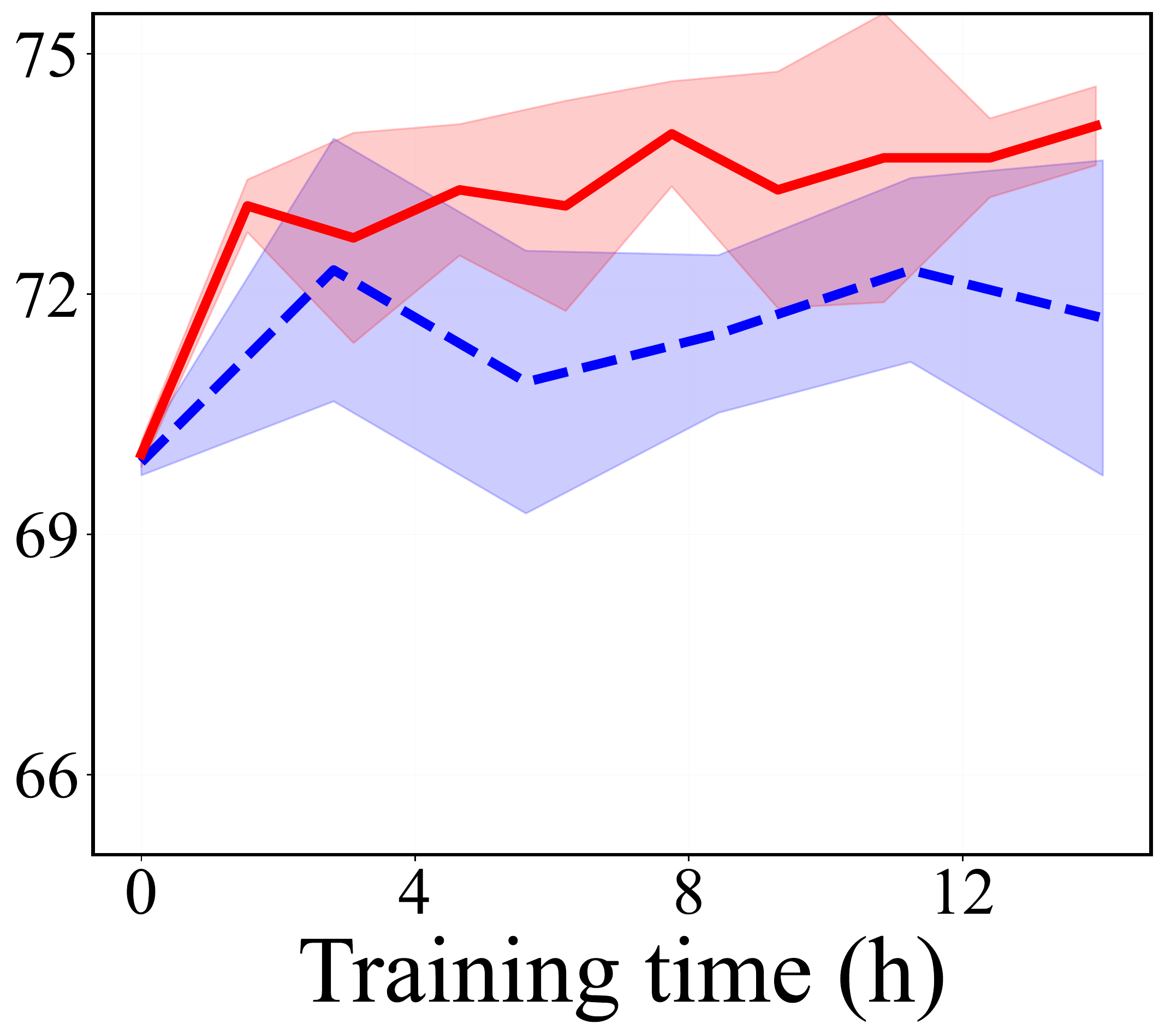}
    \caption{ImageNet-10 (Img/Cls=10)}
    \label{fig:imagenet10-10-time}
  \end{subfigure}
  \caption{Performance comparison across varying training time. Our method significantly requires less time and achieves better performance. The result reported by our method is 5 $\times$ acceleration.}
  \label{fig:time-accuracy}
  \vspace{-3mm}
\end{figure*}

\begin{figure*}[!ht]
  \centering
  \begin{subfigure}{0.235\linewidth}
    \includegraphics[width=1\linewidth]{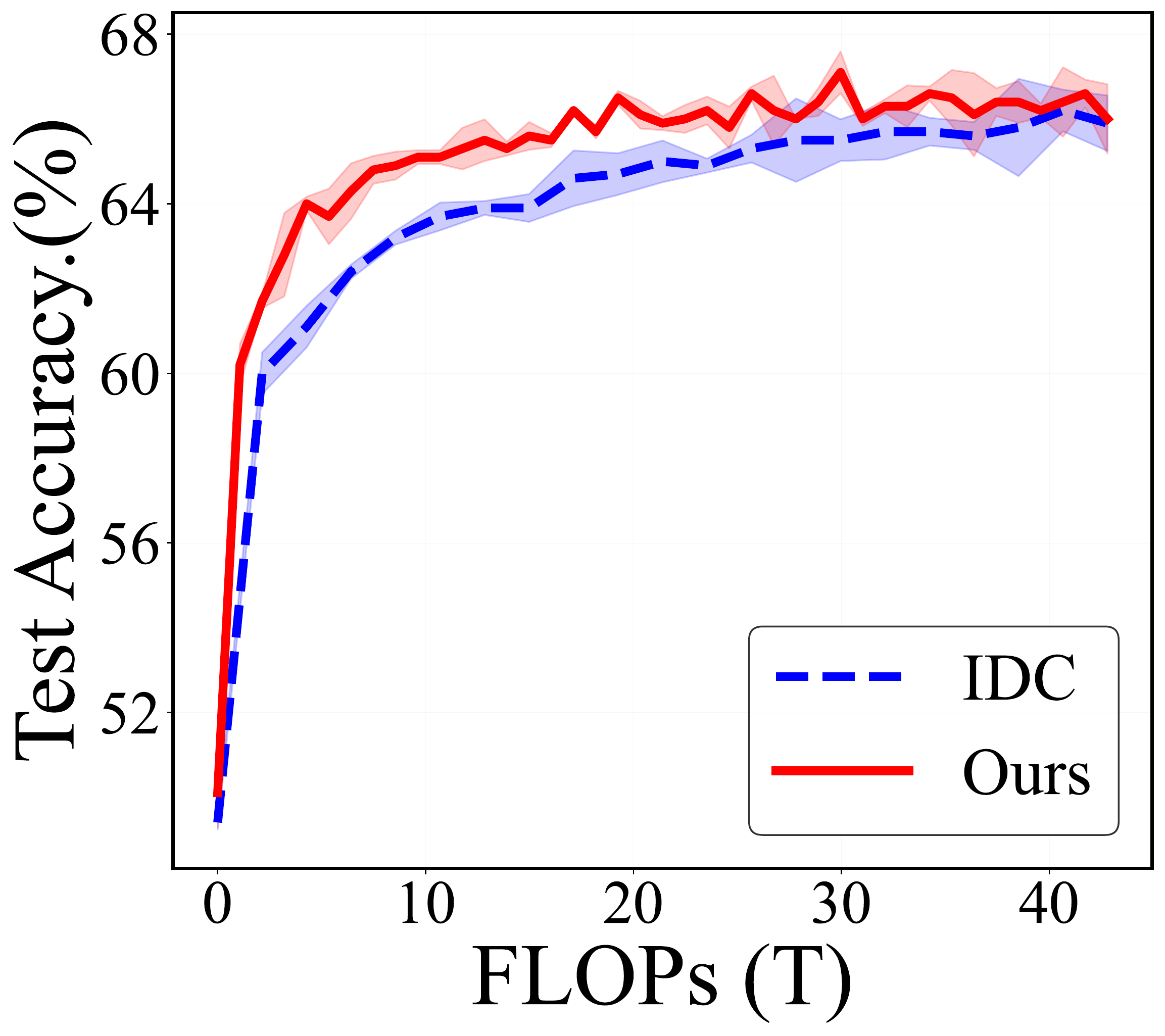}
    \caption{CIFAR-10 (Img/Cls=10).}
    \label{fig:cifar10-10-flops}
  \end{subfigure}
  \hfill
  \begin{subfigure}{0.235\linewidth}
    \includegraphics[width=1.\linewidth]{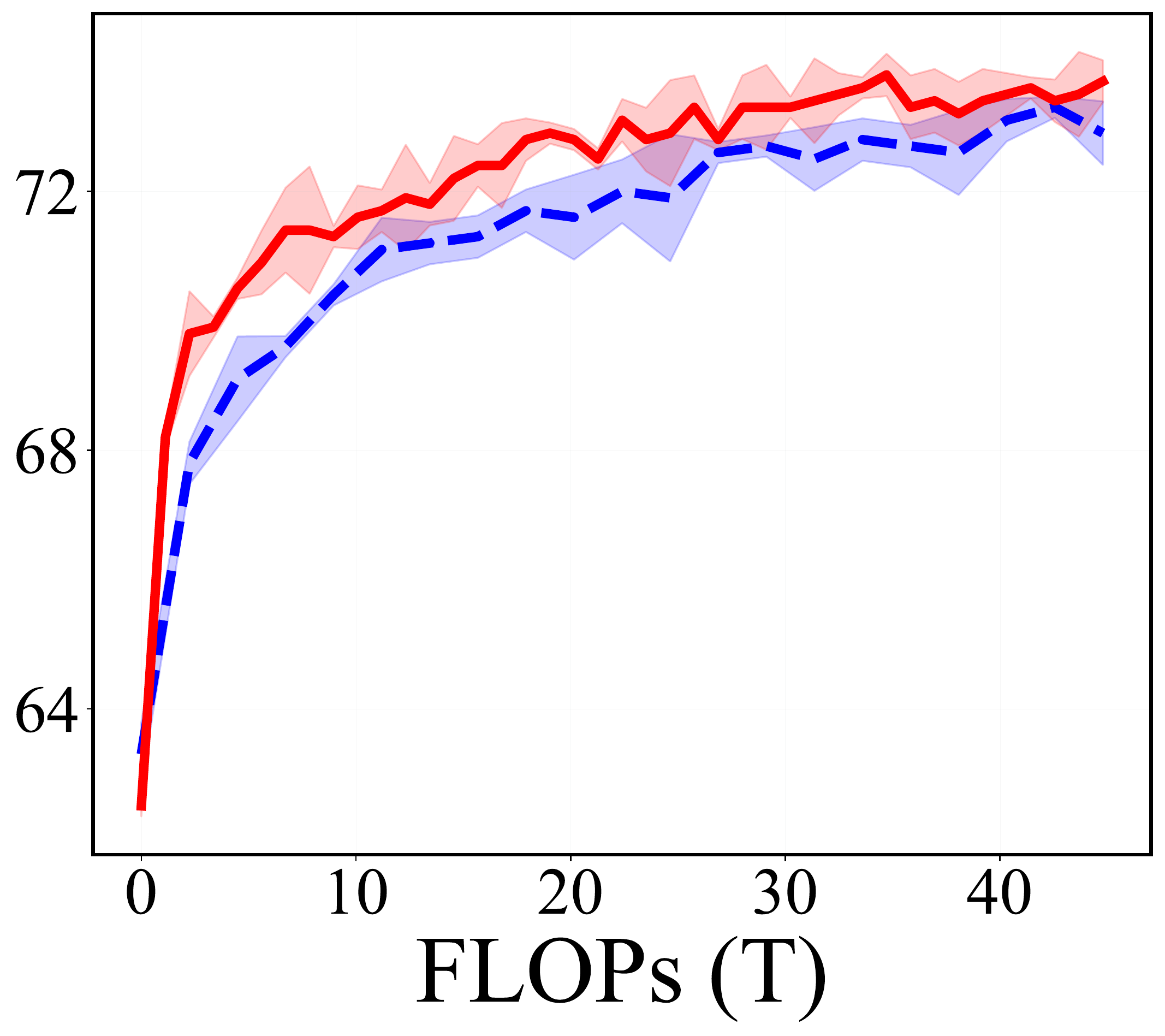}
    \caption{CIFAR-10 (Img/Cls=50)}
    \label{fig:cifar10-50-flops}
  \end{subfigure}
  \hfill
  \begin{subfigure}{0.235\linewidth}
    \includegraphics[width=1.\linewidth]{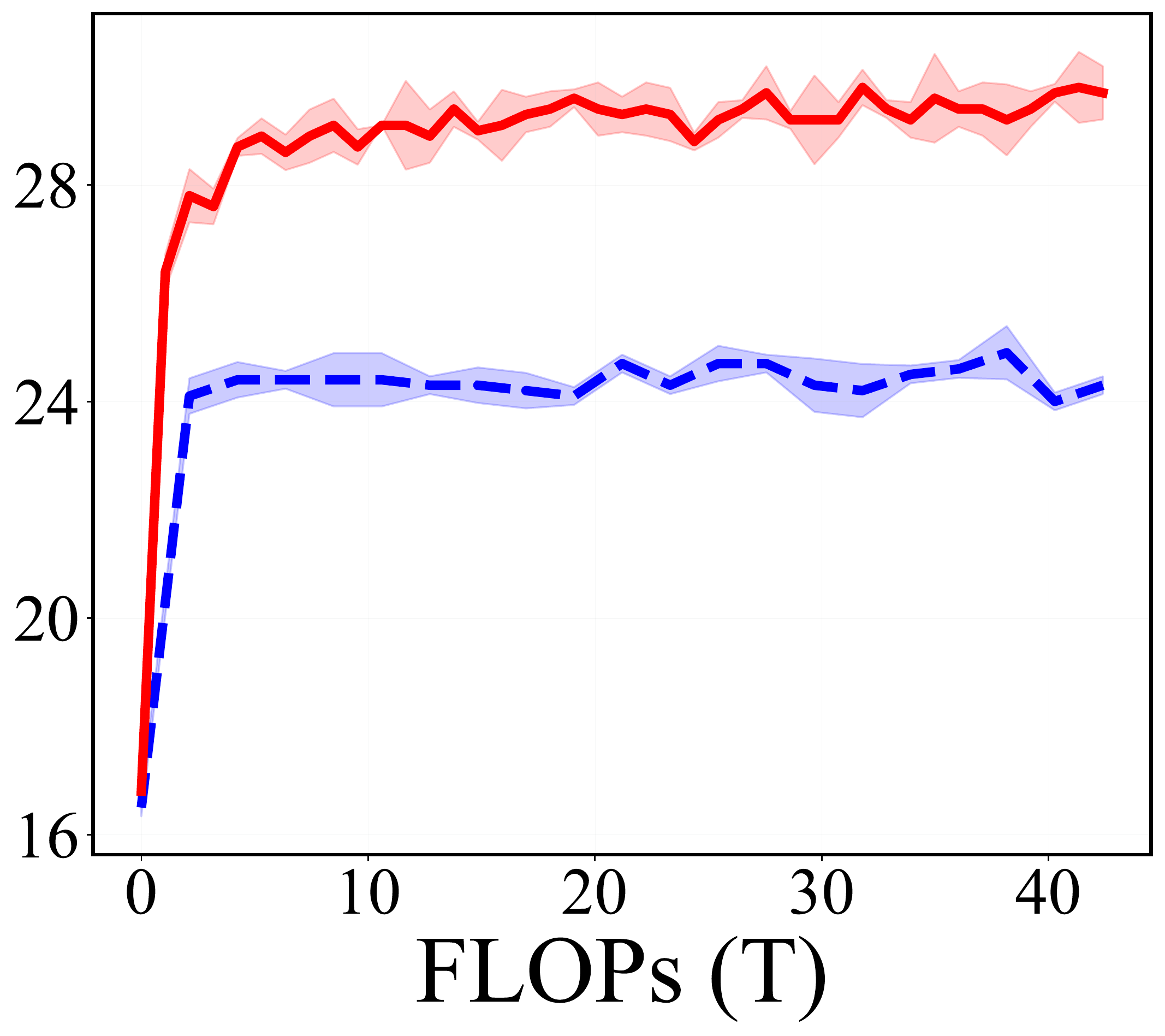}
    \caption{CIFAR-100 (Img/Cls=1)}
    \label{fig:cifar100-1-flops}
  \end{subfigure}
  \hfill
  \begin{subfigure}{0.235\linewidth}
    \includegraphics[width=1.\linewidth]{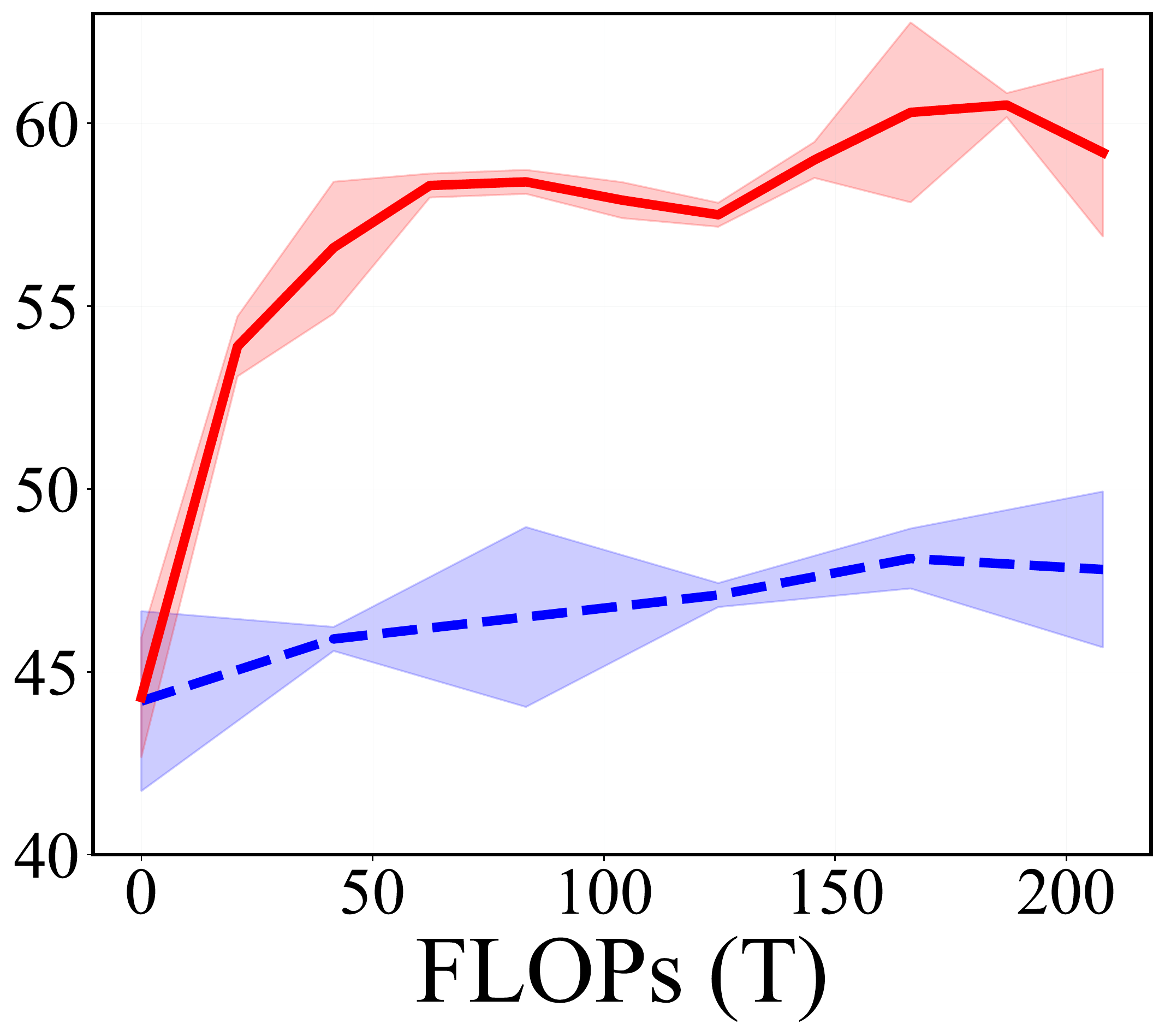}
    \caption{ImageNet-10 (Img/Cls=1)}
    \label{fig:imagenet10-1-flops}
  \end{subfigure}
  \caption{Performance comparison across varying FLOPs. Our method significantly requires less computation resource and achieves better performance. The result reported of our method is 5 $\times$ acceleration.}
  \label{fig:flops-accuracy}
  \vspace{-5mm}
\end{figure*}

\noindent \textbf{ImageNet.} Apart from CIFAR-10 and CIFAR-100, we evaluate the performance of our method on large-scale and high-resolution dataset ImageNet. As shown in \cref{tab:imagenet-add}, our method significantly outperforms all the baselines across various numbers number of classes and surpasses the leading method IDC under $5\times$ and $10\times$ speed-ups. Existing methods perform poorly on the high-resolution datasets, such as ImageNet. However, our method not only achieves a better performance but also improves the efficiency on both low- and high-resolution datasets. This demonstrates the effectiveness and scalability of our method and makes it more appealing from all practical purposes.

\subsection{Comparison on Training Budgets}

We further investigate the efficiency of our method by considering various amounts of training budgets. We consider training budgets from two perspectives, time efficiency, including training steps and training times as shown in ~\cref{fig:epoch-accuracy-add} and ~\cref{fig:time-accuracy}, and computation efficiency, including FLOPs as shown in ~\cref{fig:flops-accuracy}. We remark that our method provides significantly better performance than the leading method IDC across all ranges of budgets. Our method requires fewer training steps, consumes shorter training time, and fewer computation resources to reach comparable or better performance, which demonstrates the efficiency of our method.


\begin{table}[!ht]
    \setlength{\tabcolsep}{2pt}
    \centering
    \vspace{-3mm}
    \begin{tabular}{lccc}
    \toprule
    {Method} & {Acc.(\%) / Time(h)} & {Speed Up} & {Acc. Gain} \\
    \midrule
    {DC} & {26.0 / 1.38} & \multirow{2}{*}{$\boldsymbol{4.95\times}$} & \multirow{2}{*}{$\boldsymbol{1.11\times}$}\\
    {DC + Ours} & {\textbf{28.9 / 0.27 }}\\
    \midrule
    {DM} & {48.9 / 0.26} & \multirow{2}{*}{$\boldsymbol{4.92\times}$} & \multirow{2}{*}{$\boldsymbol{1.02\times}$}\\
    {DM + Ours} & {\textbf{49.5 / 0.05}}\\
    \midrule
    {IDC} & {\textbf{67.5} / 22.2} & \multirow{2}{*}{$\boldsymbol{4.90\times}$} & \multirow{2}{*}{$\boldsymbol{0.99\times}$}\\
    {IDC + Ours} & {67.1 / \textbf{4.45}}\\
    \bottomrule
    \end{tabular}
    \vspace{-3mm}
    \caption{\footnotesize{Applying our method to different dataset distillation methods on CIFAR-10 (10 images per class).}}
    \vspace{-4mm}
    \label{tab:dcdm}
\end{table}

\subsection{Our method with Other Algorithms}

Our weight perturbation strategy can be orthogonally applied to other dataset distillation methods. To verify the generality of our strategy, we apply the weight perturbation on other DD methods, gradient-matching DC~\cite{DBLP:conf/iclr/ZhaoMB21} and distribution-matching DM~\cite{DBLP:journals/corr/abs-2110-04181} to accelerate the training $5\times$ faster. \cref{tab:dcdm} summarizes the test performance of condensed data on CIFAR-10. The table shows that our method can also improve the performance of DC and DM.

\subsection{Model Selection Effect on Distillation}

We conduct an ablation study investigating the effect of model selection in \cref{tab:ensabla}. Instead of randomly selecting a model from pre-trained models, we consider to average the weights of all the pre-trained models at the beginning of each outer loop in our method. The table shows that the gap between random selection and weight average is no more than $1\%$ in most of datasets. It indicates that both methods preserve the relevant information of feature spaces and verifies that our method does not mainly rely on the way of model selection. 

\begin{table}[ht]
    \centering
    \begin{tabular}{cccc}
    \toprule
    {Dataset} & {Img/Cls} & {\makecell{Random\\Selection}} & {\makecell{Weight\\Average}} \\
    \midrule
    \multirow{2}{*}{CIFAR-10} & {1} & {\textbf{49.2 (0.4)}} & {48.7 (0.4)} \\
    {} & {10} & {\textbf{67.1 (0.2)}} & {66.1 (0.2)} \\
    \midrule
    \multirow{2}{*}{CIFAR-100} & {1} & {\textbf{29.8 (0.2)}} & {28.5 (0.1)}  \\
    {} & {10} & {\textbf{46.2 (0.3)}} & {43.6 (0.4)} \\
    \midrule
    \multirow{2}{*}{ImageNet-10} & {1} & {\textbf{60.5 (0.2)}} & {59.7 (1.2)} \\
    {} & {10} & {74.6 (0.3)} & {\textbf{74.8 (0.4)}} \\
    \bottomrule
    \end{tabular}
    \caption{Comparing performance of dataset distillation on different methods of method selection. The results are reported under $5\times$ acceleration with an identical training strategy.}
    \label{tab:ensabla}
\end{table}

\subsection{Visual Examples}

\begin{figure*}
  \centering
  \includegraphics[width=1\linewidth]{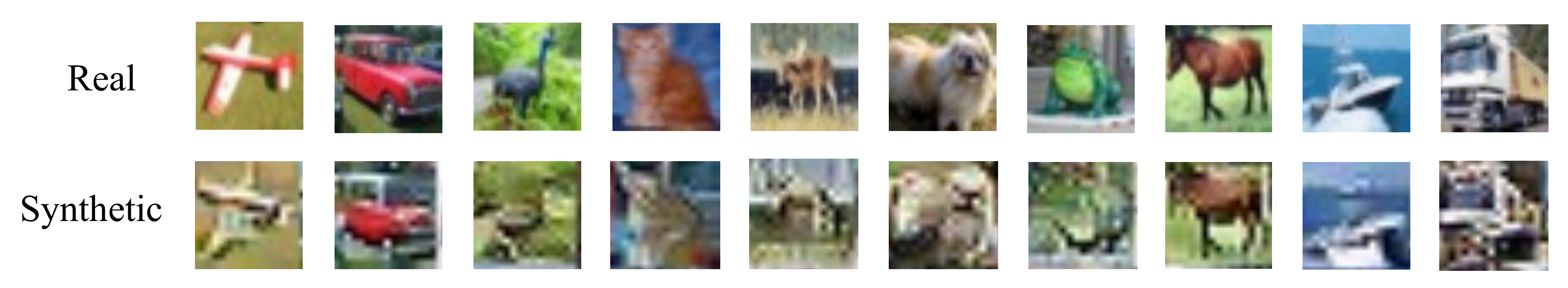}
  \caption{Comparison of real and synthetic images on CIFAR-10.}
  \label{fig:cifar10-syn}
\end{figure*}

\begin{figure*}
  \centering
  \includegraphics[width=1\linewidth]{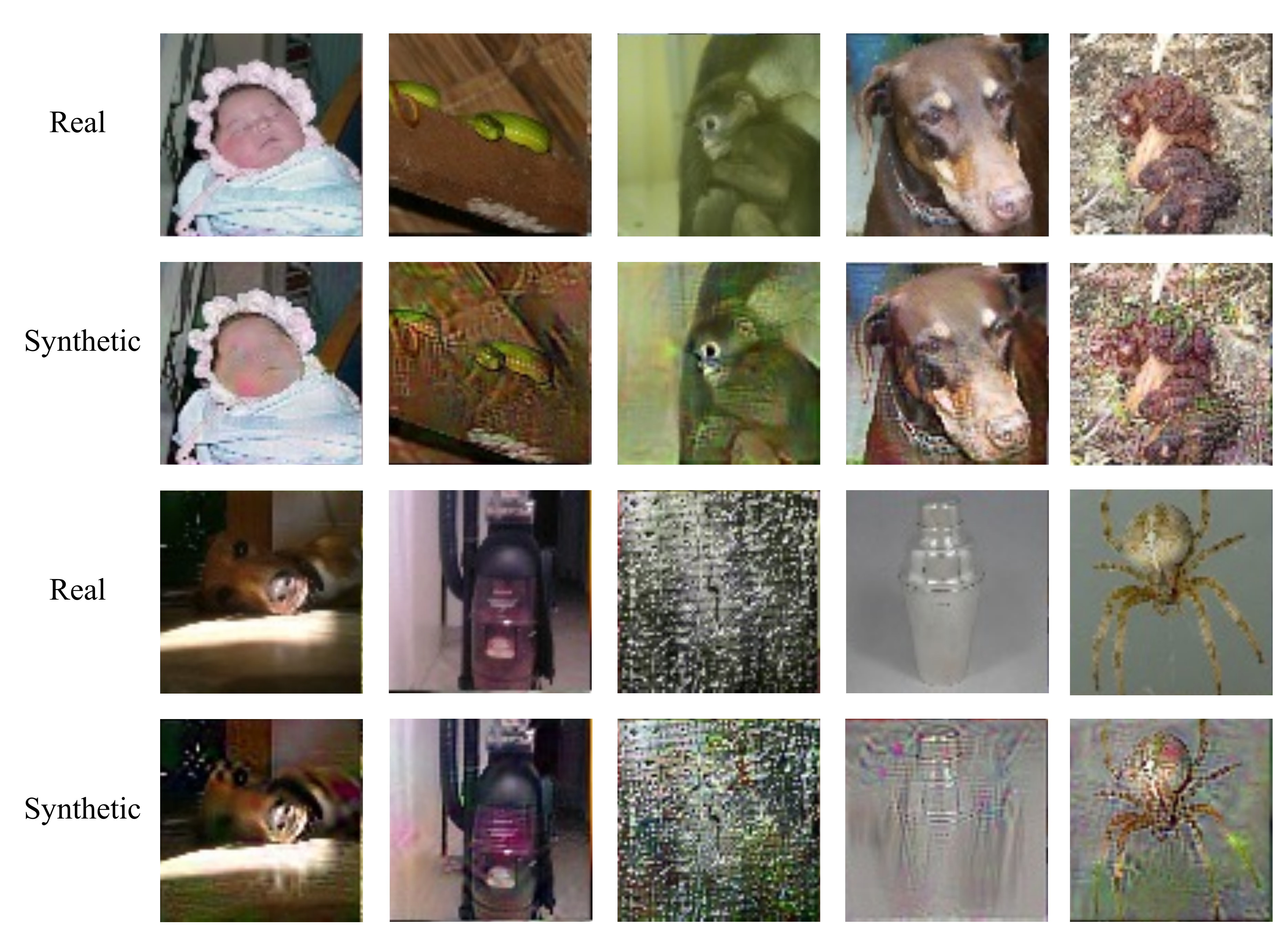}
  \caption{Comparison of real and synthetic images on ImageNet.}
  \label{fig:imagenet10-syn}
\end{figure*}

\begin{figure*}[ht]
  \centering
  \includegraphics[width=.5\linewidth]{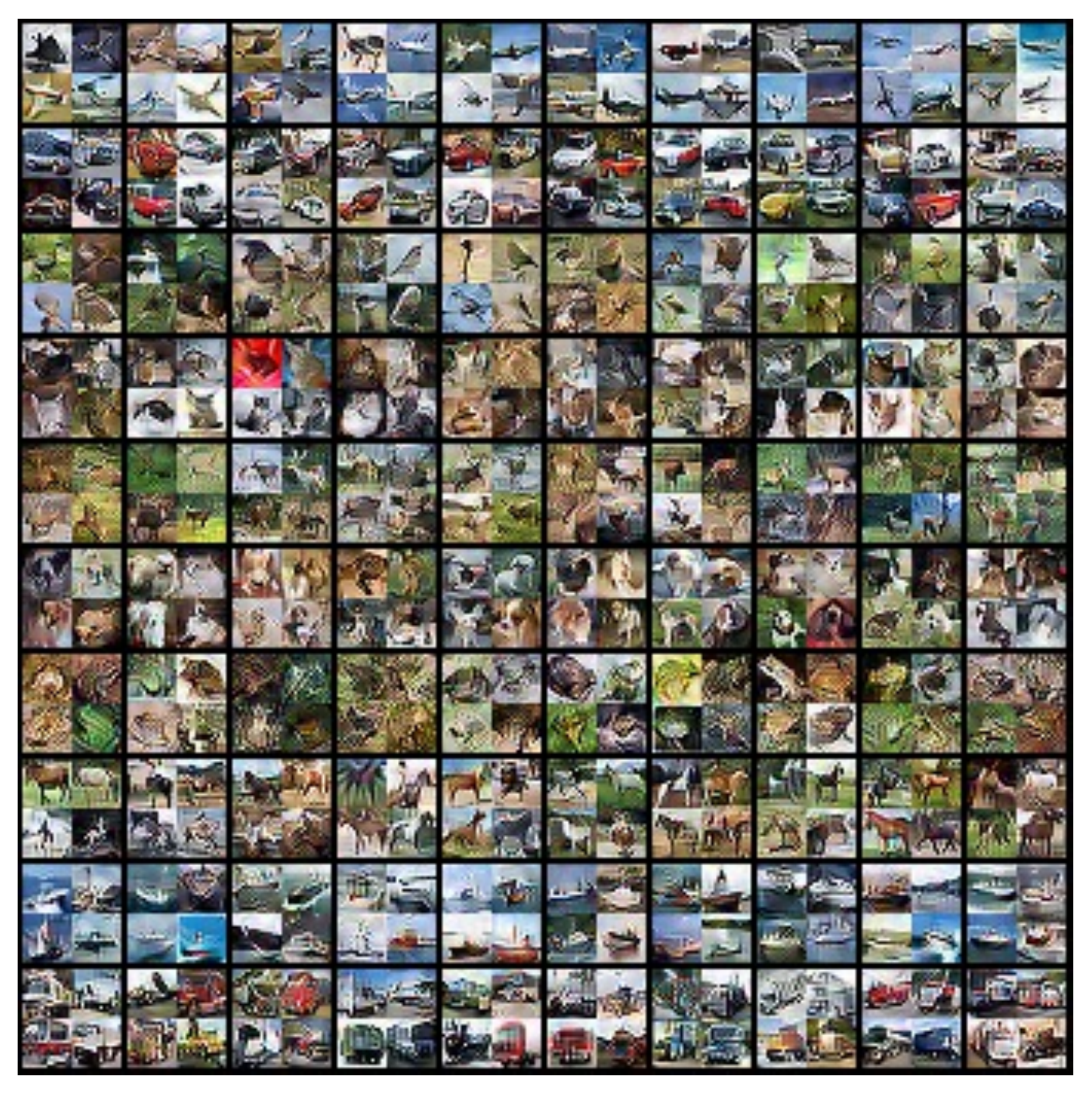}
  \caption{Condensed images of CIFAR-10 dataset 10 Img/Cls. Each row corresponds to the condensed class of a single class.}
  \label{fig:cifar10-syn-all}
\end{figure*}

\begin{figure*}[ht]
  \centering
  \includegraphics[width=.5\linewidth]{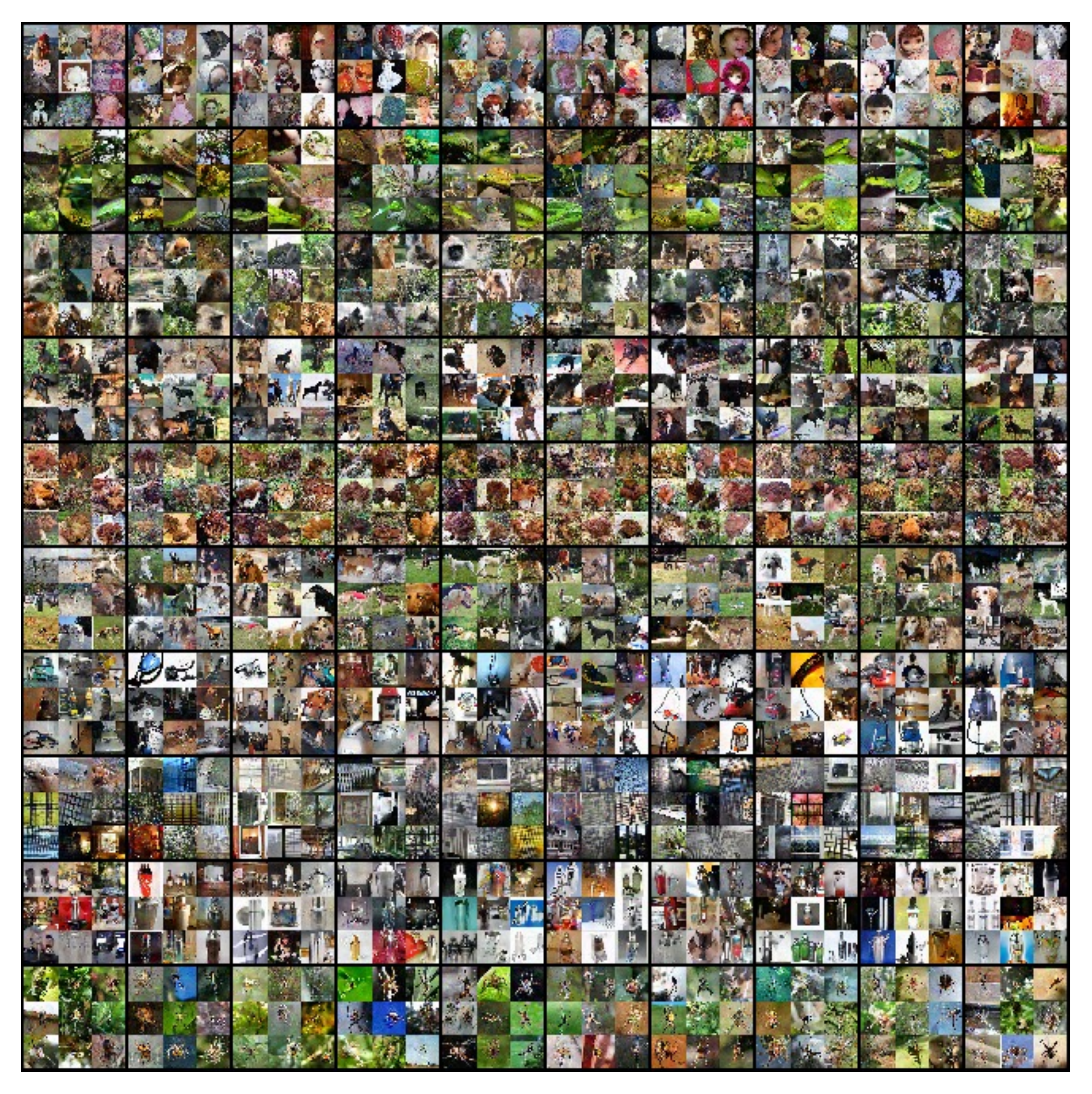}
  \caption{Condensed images of ImageNet-10 dataset 10 Img/Cls.}
  \label{fig:imagenet10-syn-all}
\end{figure*}

We provide visual examples of our method on CIFAR-10, ImageNet under $5\times$ acceleration in the following pages. In ~\cref{fig:cifar10-syn} and ~\cref{fig:imagenet10-syn}, we compare our synthetic data samples to the real training data samples, which we used as initialization of the synthetic data. From the figure, we remark that our condensed data looks abstract, yet still recognizable, representative of each class. We also provide the full condensed data in ~\cref{fig:cifar10-syn-all} and ~\cref{fig:imagenet10-syn-all}, under the storage of 10 images per class.

\newpage
{\small
\bibliographystyle{cvpr2023_conference}
\bibliography{cvpr2023_conference}
}



